\documentclass[unnumsec,webpdf,contemporary,large]{oup-authoring-template}%

\usepackage{makecell}
\usepackage{colortbl}
\usepackage{xcolor}
\usepackage{soul}
\usepackage{booktabs}
\usepackage{csquotes}
\usepackage{pdflscape}
\usepackage{adjustbox}
\usepackage{multirow}
\usepackage{pifont}
\usepackage{afterpage}
\usepackage{tabularx}
\usepackage{paralist}
\usepackage[caption=false]{subfig}

\hypersetup{
	colorlinks = true,
	citecolor = blue,
	linkcolor = blue,
	urlcolor = blue,
}

\newcommand{\cmark}{\ding{51}}

\newcommand{\refappendix}[1]{Appendix~\nameref{#1}}

\graphicspath{{Fig/}}

\theoremstyle{thmstyleone}%
\theoremstyle{thmstyletwo}%
\theoremstyle{thmstylethree}%

\begin{document}

\journaltitle{Journal Title Here}
\DOI{DOI HERE}
\copyrightyear{2023}
\pubyear{2023}
\access{Advance Access Publication Date: Day Month Year}
\appnotes{Paper}

\firstpage{1}


\title{HunFlair2 in a cross-corpus evaluation of biomedical named entity recognition and normalization tools}

\author[1,$\ast$,$\dagger$]{Mario Sänger\ORCID{0000-0002-2950-2587}}
\author[1,$\dagger$]{Samuele Garda}
\author[1,$\dagger$]{Xing David Wang}
\author[2]{Leon Weber-Genzel}
\author[1]{Pia Droop}
\author[3]{Benedikt Fuchs}
\author[1]{Alan Akbik}
\author[1,$\ast$]{Ulf Leser}

\authormark{Sänger et al.}

\address[1]{\orgdiv{Department of Computer Science}, \orgname{Humboldt-Universität zu Berlin}, \orgaddress{\street{Unter den Linden 6}, \postcode{10099 Berlin}, \country{Germany}}}

\address[2]{\orgdiv{Center for Information and Language Processing (CIS)}, \orgname{Ludwig Maximilian University Munich}, \orgaddress{\street{Geschwister-Scholl-Platz 1}, \postcode{80539 München}, \country{Germany}}}

\address[3]{\orgdiv{}, \orgname{Research Industrial Systems Engineering (RISE) Forschungs-, Entwicklungs- und Großprojektberatung GmbH}, \orgaddress{\street{Concorde Business Park F}, \postcode{2320 Schwechat}, \country{Austria}}}

\corresp[$\ast$]{Corresponding authors: \href{email:saengema@informatik.hu-berlin.de}{saengema@informatik.hu-berlin.de} and \href{email:leser@informatik.hu-berlin.de}{leser@informatik.hu-berlin.de}\\$^\dagger$Authors contributed equally.}

\received{Date}{0}{Year}
\revised{Date}{0}{Year}
\accepted{Date}{0}{Year}



\abstract{%
   With the exponential growth of the life sciences literature,
   biomedical text mining (BTM) has become an essential technology for accelerating the extraction of insights from publications.
    Entity extraction, i.e., the identification of entities in texts, such as diseases, drugs, or genes, and entity normalization, i.e., their linkage to a reference knowledge base, are crucial steps in any BTM pipeline to enable information aggregation and integration of information contained in many different documents. However, tools for these two steps are rarely applied in the same context in which they were developed. Instead, they are applied \enquote{in the wild},
    i.e., on application-dependent text collections ranging from moderately to extremely different
    from those used for the tools' training, varying, e.g., in focus, genre, style, and text type. This raises the question of whether the reported performance of BTM tools, usually obtained by training and evaluating on different partitions of the same corpus, can be trusted for downstream applications. Here, we report on the results of a carefully designed \textit{cross-corpus} benchmark for named entity extraction, where tools were applied systematically to corpora not used during their training. 
    Based on a survey of 28 published systems, we selected five, based on pre-defined criteria like feature richness and availability, for an in-depth analysis on three publicly available corpora encompassing four different entity types. 
     Comparison between tools results in a mixed picture and shows that the performance in a cross-corpus setting is significantly lower than in an in-corpus setting. 
    HunFlair2, the redesigned and extended successor of the HunFlair tool, showed the best performance on average, being closely followed by PubTator.  
    Our results indicate that users of BTM tools should expect diminishing performances when applying them in \enquote{the wild} compared to original publications and show that further research is necessary to make BTM tools more robust. 
}




\maketitle

\section{Introduction}\label{sec:intro}
\begin{figure}[t!]
\includegraphics[width=\columnwidth]{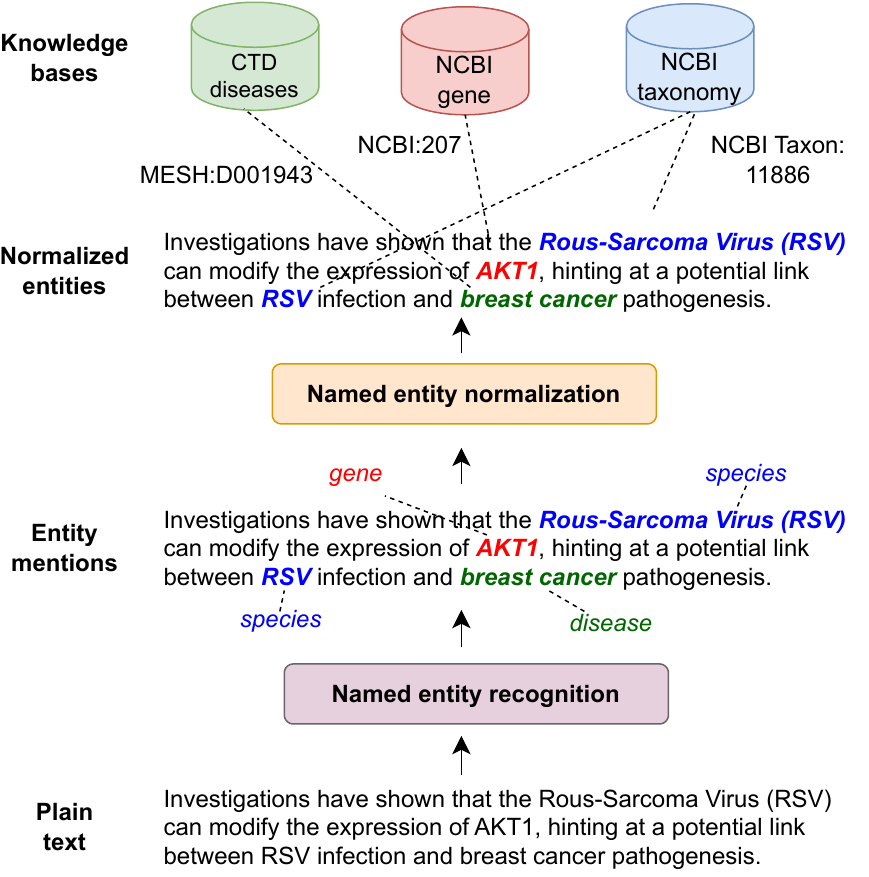}
\centering
\caption{Illustration of the named entity extraction process. First entity mentions in plain will be identified using named entity recognition (NER) tools. Afterwards named entity normalization (NEN) approaches map the found mentions to standard identifiers in a knowledge base.}
\label{fig:ee_pipeline}
\end{figure}

The volume of biomedical literature is expanding at a rapid pace, with public repositories like PubMed housing over 30 million publication abstracts.
A major challenge lies in the high-quality extraction of relevant information from this ever-growing body of literature, a task that no human can feasibly accomplish, thus requiring support from computer-assisted methods, usually in the form of information extraction pipelines. 
A crucial step in such pipelines is the extraction of biomedical entities (such as genes/proteins, diseases, or drugs/chemicals) because it is a prerequisite for further processing steps, like relation extraction~\cite{weber2022chemical}, knowledge base (KB) completion \cite{sanger2021large} or pathway curation \cite{weber2020pedl}. As shown in Figure~\ref{fig:ee_pipeline}, this step typically involves two different sub-steps: (1) named entity recognition (NER) and (2) named entity normalization (NEN)\footnote{Also known as entity linking or entity disambiguation. We refer to their combination as extraction}~\cite{huang2020biomedical}. 
NER identifies and classifies biomedical entities discussed in a given document. However, different documents may use different names (synonyms) to refer to the same biomedical concept. 
For instance, \enquote{tumor protein p53} or \enquote{tumor suppressor p53} are both \textit{valid} names for the gene \enquote{TP53} (NCBI Gene: 7157).
Moreover, the same entity mention could refer to multiple different concepts (homonyms), e.g., \enquote{RSV} could be \enquote{Rous-Sarcoma-Virus} or \enquote{Respiratory syncytial virus} depending on the context. 
Entity normalization addresses the issues of synonyms and homonyms and maps the mentions found by NER to a standard form, usually a knowledge base (KB) identifier. 
This process ensures that all mentions of an entity, regardless of how they are expressed in the text, are recognized as referring to the same real-world object, making it easier to aggregate, compare, and integrate information across different documents.

Over the last two decades, several studies investigated biomedical named entity recognition and normalization \cite{huang2020biomedical,SongLLZ21,wang2023pre}.
%
%
Of the many research prototypes, some have been consolidated into mature and easy-to-install or use \textit{tools} that end users can apply directly for their specific needs \cite{weiPubTatorCentralAutomated2019,weber2021hunflair,zhang_biomedical_2021}. 
However, this involves deploying them \enquote{in the wild}, i.e., to specifically created text collections with arbitrary focused topics (e.g., cancer or rare genetic disorders), entity distribution (gene-focused molecular biology or disease-focused clinical trials), genre (publication, patent, report) and text type (abstract, full text, user-generated content). But the tools originally were trained and evaluated on a single or few gold standard corpora, which each was specially created for a biomedical sub-domain (e.g., cancer \cite{pyysalo2013overview} or plants \cite{kim2019corpus} a specific text type and genre, and with a specific entity distribution. 
The mismatch between these two settings, i.e., training/evaluation vs downstream deployment, raises the question of whether the performance in the first can be trusted to estimate the one achievable in the second.
As named entity extraction is the cornerstone of several applications, e.g., for relation and event extraction \cite{LeeYKKKSK20,trieuDeepEventMineEndtoendNeural2020,wang-etal-2020-biomedical}, the issue has a direct and critical impact on downstream information extraction pipelines. 
For instance, \cite{li2023rethinking} investigate the robustness of
document-level relation extraction models.
They discover that systems experience a sharp performance drop if the entities mentioned are not identified correctly, illustrating the vulnerability of existing state-of-the-art relation extraction models to cascading errors.

To better quantify the impact of this issue, previous work proposed to use \textit{cross-corpus} evaluation, i.e., training models on one corpus and evaluating on a different one \cite{ExploitingAndGalea2018, giorgi2020towards,weber2020huner,weber2021hunflair}. 
For instance, Giorgi and Bader \cite{giorgi2020towards} show that the performance of neural networks for NER drops by an average of 31.16\% F1 when tested on a corpus different from the one used for training. 
Previous studies, however, present a few limitations: 
First, existing benchmarking studies, both in- and cross-corpus, focus primarily either on recognition \cite{song2021deep,huang2020biomedical,su2022deep} or normalization \cite{AnOverviewOfFrench2022,garda2023belb,AComprehensiveKartch2023} but do not provide results for named entity extraction, i.e., end-to-end NER and NEN.
Secondly, many of them do not account for the latest technologies and models
\cite{wang2018comparative,alshaikhdeeb2016biomedical} like transformer-based language models \cite{Bern2AnAdvanMujeen}. 

In this study, we address these limitations and present the first cross-corpus evaluation of state-of-the-art (SOTA) tools for named entity extraction (end-to-end NER and NEN) in BTM to provide an in-depth analysis of their results that highlights current limits and areas for improvement. 
Based on an extensive literature review identifying 28 biomedical text mining models, we selected five mature tools based on predefined criteria: The tool must (C1) support both NER and NEN, (C2) integrate the most recent improvements from the machine-learning / NLP community, (C3) extract at least the most common biomedical entity types (genes, disease, chemicals, and species), and (C4) require no further licenses (e.g., UMLS license). Four tools quailfied, namely BERN2~\cite{Bern2AnAdvanMujeen}, PubTator~\cite{weiPubTatorCentralAutomated2019}, SciSpacy~\cite{neumann2019scispacy}, and bent~\cite{ruas2023lasige}. As a fifth tool, also fulfilling these criteria, we present HunFlair2, our novel and extended version of HunFlair \cite{weber2021hunflair}, which we integrated directly into the NLP framework flair\footnote{\url{https://github.com/flairNLP/flair}}. 

We performed extensive experiments\footnote{Code to reproduce our results is available at: \url{https://github.com/hu-ner/hunflair2-experiments}} with all five tools on three corpora covering four entity types and diverse forms of text (i.e., scientific abstracts, full texts, and figure captions). The corpora were explicitly selected such that none of the tools (according to their documentation) ever used them for training. Our results show stark performance variations among tools w.r.t. entity types and corpus, with differences of up to 46 pp when comparing scores to those of previously published in-corpus results. The overall most robust tool was HunFlair2, reaching an average F1 score of 58,79\% and the best results in two (chemicals and diseases) of four entity types. PubTator scores very close second best on average and best in the other two types (genes and species).
This highlights the need for further research on named entity extraction tools to increase their robustness towards changes in biomedical subdomains and text types.

\section{Background}
Named entity extraction tools usually decompose the task into NER and NEN, so we provide an overview of existing approaches for both subtasks in this section.

\subsection{Biomedical named entity recognition}
%
Traditional methods for NER used dictionary-based and rule-based techniques \cite{AkhondiHHMK15}.
The former involves creating dictionaries composed of extensive lists of names
that serve as a reference for individual entity types, typically combined with some form of fuzzy text matching.
The effectiveness of these approaches heavily depends on (i) the dictionary's quality, its size and range of vocabulary, and (ii) the type of matching strategy between dictionary entries and candidate entities.
Rule-based approaches instead identify concepts by employing a set of handcrafted or automatically learned rules derived from
textual or syntactical patterns identified in scientific literature.
Despite these methods often delivering high precision, both struggle with difficult maintenance and extensibility, out-of-vocabulary terms, and unusual contexts, usually resulting in low recall \cite{huang2020biomedical}.
%

In response to these limitations, researchers in the last two decades turned to machine learning (ML) as base technology to train a model capable of recognizing similar patterns in unseen data.
First approaches for ML-based NER, such as Support Vector Machines (SVM) \cite{KazamaMOT02}, Hidden Markov Models (HMM) \cite{ShenZZST03}, and Conditional Random Fields (CRF) \cite{Settles04}, used handcrafted features to represent tokens.
Within these methods, NER is framed as a sequence labeling problem, i.e., each token in a given sentence is assigned a label representing whether the token constitutes a part of a named entity and, if so, the specific entity type it belongs to.

However, feature engineering is labor-intensive and time-consuming and ignores the semantic similarity between different tokens. Accordingly, the next generation of NER tools applied representation learning and word embeddings \cite{HabibiWNWL17}. 
In representation learning, models automatically learn optimal features for the given task without human intervention \cite{SongLLZ21}. Several studies explored different neural architectures for this purpose, e.g., convolutional \cite{KorvigoHZS18,ChoHPP20} and recurrent \cite{JuMA18,DangLNV18,LyuCRJ17} ones, with bidirectional LSTMs (BiLSTM) \cite{hochreiter1997long} showing the best performance. Word embeddings refer to techniques representing tokens as (context-dependent) low-dimensional vectors learned from large, domain-specific corpora in a pre-processing step. Popular models include BioBERT \cite{LeeYKKKSK20}, BlueBERT \cite{PengYL19}, PubMedBERT \cite{GuTCLULNGP22}, and BioLinkBERT \cite{YasunagaLL22}, with the latter currently achieving the best performance in general BTM.

More recently, two additional techniques were found to be widely adopted to improve identification accuracy. First, many tools started to use multiple corpora in the training process to improve model generalization \cite{giorgi2020towards,weber2021hunflair,luo2023aioner}. For instance, HunFlair \cite{weber2021hunflair} is trained on 23 corpora processed in a unified format and performs better than single-corpus models in a cross-corpus evaluation.
Second, recent tools trained joined models for different entity types instead of learning an individual model for each type. For instance, AIONER (All-In-One NER) \cite{luo2023aioner} supports the simultaneous identification of different entity types by prepending special task tokens to the input text (e.g., \textit{$<$gene$>$} for genes), thus enabling the inclusion of only partially annotated corpora (e.g., only disease and chemicals).

\subsection{Biomedical named entity normalization}


Traditional techniques for entity normalization perform entity recognition and
normalization in one step~\cite{leser2005makes}. They use different forms of string matching or
dictionary lookup, often combined with handcrafted decision rules to address
specific needs of the targeted entity type, e.g., GNormPlus \cite{Wei2015} for
genes and tmVar \cite{Wei2022} for variants.

Later approaches are based on independent models for the two tasks to better cope with their largely varying characteristics. For instance, while NER must choose only from a few different classes for each token during sequence labeling, NEN sometimes has to consider hundreds of thousands of candidates for a given mention. Focusing on the NEN step alone, researchers also in NEN switched to machine learning-based approaches
\cite{kaewphan2018wide,cho2017method,Leaman2016}. For instance,
\cite{leaman2013dnorm} addresses disease normalization by modeling the problem
as a pairwise learning-to-rank task, while \cite{Leaman2016} introduces
TaggerOne, a semi-Markov model, that learns to perform entity recognition and
normalization jointly. With the advent of DL, research explored
neural network architecture and training strategies as done in NER. The main idea of these
approaches is to encode surface forms of entities, originating either from KBs
or mentions in free text, into a common embedding space and leveraging concept
similarity within that latent space for linking. For instance \cite{Phan2019}
train a character-based BiLSTM with multiple objective functions to maximize
the similarity of a name in a KB to its mentions and context. Similarly,
\cite{Chen2021} uses a CNN to obtain embeddings and introduce a coherence loss, 
penalizing the model if linked entities in a document are unlikely to co-occur.

Most recent studies have developed linking methods based on pre-trained
language models. One prominent example is BioSyn \cite{Sung2020}, which uses the BioBERT language model as the encoder. During training, given a mention, its entity, and a
candidate pool, it learns to maximize the similarity between the mention
and all names whose associated entity is the same as the gold one.
\cite{BiomedicalConcYanC2021} proposed to enhance BioSyn by
leveraging hierarchical information contained in KBs, i.e., hypernym and hyponym
relations. A primary challenge of biomedical NEN is data scarcity, which is a
major hindrance for DL-based models. To address this challenge, researchers
have explored different strategies for targeted pre-training of language models
for NEN, involving the usage of KB data \cite{Liu2021a} or weakly labeled
corpora from sources such as Wikipedia or PubMed
\cite{CrossDomainDaVarma2021,zhang2022knowledge}. Other relevant research
directions are clustering-based approaches \cite{Angell2021,Agarwal2022} and
generative models \cite{GenerativeBiomYuan2022}. The latter tackles
normalization as a sequence-to-sequence task and trains models that generate a
valid (preferred) entity name of the KB under consideration.

All approaches mentioned so far focus solely on entity normalization, i.e., they
assume that mentions have already been recognized. In contrast,
\cite{EndToEndBiomUjiie2021} and \cite{Bhowmik2021} investigate methods to
solve NER and NEN jointly using pre-trained model language models and combining
it with string matching, respectively. Furthermore, most research in DL-based methods addresses only two entity types, namely diseases and chemicals. Applying them to other types, especially genes or species, leads to significant performance drops \cite{BelbABiomediGarda2023}. This can be attributed to the steep growth in frequencies of homonyms and synonyms, increasing the importance of context for correct normalization. Finally, DL-based methods face scaling issues, i.e., they cannot handle KBs with tens of millions of entries, as required, for instance, by genes. Due to these restrictions, even today, rule-based (or hybrid)  approaches
still dominate in state-of-the-art tools
\cite{weiPubTatorCentralAutomated2019,Bern2AnAdvanMujeen}.

\section{Biomedical entity extraction tools}\label{sec:bio_ee_tools}

\begin{table*}[htbp]
  \centering
  \caption{%
  Overview of the tools selected for our evaluation.
  We distinguish rule-based (\enquote{RB}), machine learning-based (\enquote{ML}) and neural-network based (\enquote{NN}) approaches for NER and NEN.
  Moreover, for each tool we illustrate the support of the following entity types: genes (Ge), species (Sp), disease (Di), chemical (Ch), cell line (Cl) and variant (Va).
  For each entity type we illustrate whether the tool supports NER and NEN of the type by marking the column with \cmark, if only NER is supported we use (\cmark).
  Last update highlights the last update of the code repository of the respective tool.
  Citations counts are taken from Google Scholar on 01/10/2024. 
  }
    \begin{tabular}{lcccccccccccc}
        \toprule
    \multicolumn{1}{c}{\textbf{Tool}} &
      \multicolumn{1}{c}{\textbf{API}} &
      \multicolumn{1}{c}{\textbf{Ge}} &
      \multicolumn{1}{c}{\textbf{Sp}} &
      \multicolumn{1}{c}{\textbf{Di}} &
      \multicolumn{1}{c}{\textbf{Ch}} &
      \multicolumn{1}{c}{\textbf{Cl}} &
      \multicolumn{1}{c}{\textbf{Va}} &
      \multicolumn{1}{c}{\textbf{NER}} &
      \multicolumn{1}{c}{\textbf{NEN}} &
      \multicolumn{1}{c}{\textbf{Pub. Year}} &
      \multicolumn{1}{c}{\textbf{Last Update}} &
      \multicolumn{1}{c}{\textbf{Citations}}
      \\
    \midrule
    PubTator Central &
      \multicolumn{1}{c}{REST/} &
      \multicolumn{1}{c}{\cmark} &
      \multicolumn{1}{c}{\cmark} &
      \multicolumn{1}{c}{\cmark} &
      \multicolumn{1}{c}{\cmark} &
      \multicolumn{1}{c}{\cmark} &
      \multicolumn{1}{c}{\cmark} &
      \multicolumn{1}{c}{ML / NN} &
      \multicolumn{1}{c}{RB} &
      \multicolumn{1}{c}{2019} &
      \multicolumn{1}{c}{-} &
      \multicolumn{1}{c}{315}
      \\
    \cite{weiPubTatorCentralAutomated2019} &
       Tools&
       &
       &
       &
       &
       &
       &
       &
       &
       &
       &
      
      \\
        \midrule
    BERN2 &
      \multicolumn{1}{c}{REST/} &
      \multicolumn{1}{c}{\cmark} &
      \multicolumn{1}{c}{\cmark} &
      \multicolumn{1}{c}{\cmark} &
      \multicolumn{1}{c}{\cmark} &
      \multicolumn{1}{c}{\cmark} &
      \multicolumn{1}{c}{\cmark} &
      \multicolumn{1}{c}{NN} &
      \multicolumn{1}{c}{RB / NN} &
      \multicolumn{1}{c}{2022} &
      \multicolumn{1}{c}{11/2023} &
      \multicolumn{1}{c}{46}
      \\
    \cite{Bern2AnAdvanMujeen} &
       Python&
       &
       &
       &
       &
       &
       &
       &
       &
       &
       &
      
      \\
                \midrule
    SciSpacy &
      \multicolumn{1}{c}{Python} &
      \multicolumn{1}{c}{(\cmark)} &
      \multicolumn{1}{c}{\cmark} &
      \multicolumn{1}{c}{\cmark} &
      \multicolumn{1}{c}{\cmark} &
      \multicolumn{1}{c}{\cmark} &
      \multicolumn{1}{c}{(\cmark)} &
      \multicolumn{1}{c}{NN} &
      \multicolumn{1}{c}{RB} &
      \multicolumn{1}{c}{2019} &
      \multicolumn{1}{c}{10/2023} &
      \multicolumn{1}{c}{635}
      \\
    \cite{neumann2019scispacy}  &
       &
       &
       &
       &
       &
       &
       &
       &
       &
       &
       &
      
      \\
        \midrule
    bent &
      \multicolumn{1}{c}{Python} &
      \multicolumn{1}{c}{\cmark} &
      \multicolumn{1}{c}{\cmark} &
      \multicolumn{1}{c}{\cmark} &
      \multicolumn{1}{c}{\cmark} &
      \multicolumn{1}{c}{\cmark} &
      \multicolumn{1}{c}{(\cmark)} &
      \multicolumn{1}{c}{NN} &
      \multicolumn{1}{c}{RB} &
      \multicolumn{1}{c}{2020} &
      \multicolumn{1}{c}{12/2023} &
      \multicolumn{1}{c}{13}
      \\
    \cite{LinkingChemicaRuas2020,ruas2023lasige} &
       &
       &
       &
       &
       &
       &
       &
       &
       &
       &
       &
      
      \\
        \midrule
    HunFlair2 &
      \multicolumn{1}{c}{Python} &
      \multicolumn{1}{c}{\cmark} &
      \multicolumn{1}{c}{\cmark} &
      \multicolumn{1}{c}{\cmark} &
      \multicolumn{1}{c}{\cmark} &
      \multicolumn{1}{c}{\cmark} &
       &
      \multicolumn{1}{c}{NN} &
      \multicolumn{1}{c}{RB / NN} &
      \multicolumn{1}{c}{2021} &
      \multicolumn{1}{c}{01/2024} &
      \multicolumn{1}{c}{83}
      \\
    \cite{weber2021hunflair} &
       &
       &
       &
       &
       &
       &
       &
       &
       &
       &
       &
      
      \\
        \bottomrule
    \end{tabular}%
  \label{tab:selected_tools}%
\end{table*}%

We now describe our process to identify and select the tools to be evaluated in our cross-corpus experiments. For the scope of our paper, we define a tool as a piece of software that is either directly accessible online via a web API or easily installable locally, e.g., by providing downloadable pre-trained models. A tool is easily usable by non-experts and provides off-the-shelf usage as far as possible without requiring extensive manual configurations. We note that this excludes research prototypes which require extra effort to be applied to new input text. 
To find such tools, we performed an extensive literature review identifying 28 biomedical entity extraction tools of largely varying functionalities and scopes. A complete list including essential characteristics can be found in Appendix~\ref{app:tools}. 
This list was further reduced to five according to the following criteria. 
\begin{enumerate}[C1: ]
	\item \label{crit:entext} A tool support both NER and NEN.
        \item \label{crit:ml_based} A tool must use machine-learning-based models (at least for NER).
        \item \label{crit:entity} A tool must extract at least genes, diseases, chemicals, and species.
        \item \label{crit:license} A tool must not require additional licenses for application (e.g., commercial or UMLS license).
	
\end{enumerate}

The rationale for \textcolor{blue}{C}\ref{crit:entext} is self-evident: we are interested in the cross-corpus performance of \textit{named entity extraction}, i.e., end-to-end NER and NEN. \textcolor{blue}{C}\ref{crit:ml_based} narrows down the scope to tools employing machine-learning-based NER as the state-of-the-art approach \cite{song2021deep}. This facilitates experiment design and cross-corpus evaluation by reducing potential overlaps of training corpora with evaluation corpora (see \nameref{sec:eval_framework} for details). \textcolor{blue}{C}\ref{crit:entity} ensures that the tool can extract the entities most important for downstream applications. We note that we do not count the variant entity type as a hard constraint since, for this entity type, to the best of our knowledge, the only existing approach is tmVar \cite{Wei2022}. 
Finally, as BTM tools are often used in research pipelines, we require that there should be no constraint on its usability/license agreements (\textcolor{blue}{C}\ref{crit:license}).

Using the criteria above, we selected the following tools for our cross-corpus evaluation:
BERN2~\cite{Bern2AnAdvanMujeen}, bent~\cite{LinkingChemicaRuas2020,ruas2023lasige}, PubTator~\cite{weiPubTatorCentralAutomated2019}, SciSpacy~\cite{neumann2019scispacy}, and HunFlair2, our updated version of HunFlair~\cite{weber2021hunflair} which is first published with this paper. Table~\ref{tab:selected_tools} provides an overview of the main features of the selected tools.
Further details on compiling the initial list of tools, their main aspects, and unfulfilled criteria can be found in Appendix~\ref{app:tools}. 

\medskip

\noindent \textbf{BERN2} \cite{Bern2AnAdvanMujeen} uses a pipeline approach for entity extraction, separating NER and NEN into two distinct steps built upon each other.
The NER models of BERN2 consist of a transformer-based RoBERTa model and a Conditional Random Field based on tmVar 2.0~\cite{wei2018tmvar}. 
The CRF is used to extract variant entities, whereas the RoBERTa model extracts the other five entity types.
The RoBERTa \cite{liu2019roberta} model is trained in a multi-task fashion to share model parameters during the training of each entity type. 
For training the NER model, \cite{liu2019roberta} use BC2GM \cite{smith2008overview} for genes, BC4CHEMD \cite{krallinger2015chemdner} for chemicals, NCBI Disease \cite{Dogan2014} for diseases, Linnaeus \cite{gerner2010linnaeus} for species and JNLPBA \cite{collier2004introduction} for cell line
%
The normalization component of BERN2 is a hybrid system that relies on a mixture of rule-based and neural-based approaches. 
BERN2 first tries to normalize entity mentions with rule-based models: GNormPlus \cite{Wei2015} for genes, a sieve-based approach for diseases \cite{SieveBasedEntDsouz2015}, tmChem\footnote{without abbreviation resolution by Ab3P \cite{AbbreviationDeSohn2008}} \cite{leaman2015tmchem} for drug/chemical and dictionary lookup for species. 
If the rule-based system fails, mentions are passed to a BioSyn model \cite{Sung2020} for genes, diseases, and chemicals. 
The BC2GN corpus \cite{smith2008overview} is used for training the gene normalization, BC5CDR for training chemical and disease normalization, and NCBI-Disease also for disease normalization.


\medskip

\noindent \textbf{bent} \cite{ruas2023lasige} also uses a pipeline approach separating NER and NEN into two steps.
The NER component comprises standard transformer-based models initialized with weights from PubMedBERT~\cite{GuTCLULNGP22}. 
The transformers are fine-tuned separately for each entity type.
Bent is trained on 15 distinct corpora for chemicals recognition\footnote{\url{https://huggingface.co/pruas/BENT-PubMedBERT-NER-Chemical}} (BC5CDR \cite{Li2016a} and NLM-Chem\cite{islamaj2021nlm}  among them), 9 corpora for disease recognition\footnote{\url{https://huggingface.co/pruas/BENT-PubMedBERT-NER-Disease}} (BC5CDR and NCBI-disease \cite{Dogan2014} among them), 19 corpora for genes\footnote{\url{https://huggingface.co/pruas/BENT-PubMedBERT-NER-Gene}} (BC2GM \cite{smith2008overview} and CRAFT \cite{cohen2017colorado} among them) and 9 corpora for species\footnote{\url{https://huggingface.co/pruas/BENT-PubMedBERT-NER-Organism}} (Linnaeus \cite{gerner2010linnaeus} and CRAFT among them). The NEN component is based on the PageRank algorithm~\cite{Page1999ThePC}. It builds a graph where nodes are names in the KB. Nodes are connected if a relation exists between them. The relations are either (i) specified in the KB (e.g., in UMLS) or (ii) automatically extracted from the training corpora. To perform mention normalization, Bent selects the node (KB name) which maximizes the coherence of the graph~\cite{LinkingChemicaRuas2020}.

%

\medskip

\noindent \textbf{PubTator} is a web-based interface providing pre-computed entity annotations for all PubMed and PubMed Central documents and an API for entity extraction on custom documents. As we have no direct access to the PubTator source code and no documentation of how often updates have been applied to the public service since the last publication, we must rely on information from several publications to describe the models used for entity extraction. 
However, the authors regularly release new publications describing
developments to the entity extraction systems so we may provide an overview of the models used according to the current publication records.
We access PubTator through its API to process raw text\footnote{\url{https://www.ncbi.nlm.nih.gov/research/pubtator/api.html} - accessed on the 2023/07/21}, i.e., we submit plain text from the corpora to be annotated.
As publications about PubTator are separated by their extracted entity type, we also report NER and NEN models for each entity type separately here.
Generally, the NER and NEN systems make use of a pipelined
approach separating recognition and normalization into two distinct steps.
Authors of \cite{Islamaj2021} report that PubTator uses an updated version of GNormPlus \cite{Wei2015}, which uses the BlueBERT language model \cite{TransferLearniPeng2019} for NER trained on the GnormPlus corpus \cite{Wei2015} and NLM-Gene \cite{Islamaj2021}.
The gene normalization component is a statistical inference network \cite{CrossSpeciesGWeiC2011} based on TF-IDF frequencies.
SR4GN (Species Recognition for Gene Normalization) \cite{Sr4gnASpecieWeiC2012} is used for species recognition and normalization, a rule-based system which, as the name suggests, is mainly a support component for gene normalization.
According to \cite{NlmChemANewIslama2021}, PubTator currently uses BlueBERT trained on BC5CDR \cite{Li2016a} and NLM-Chem \cite{NlmChemANewIslama2021} for chemical NER, and a multi-terminology candidate resolution algorithm (MTCR) for chemical
normalization, which employs multiple string-matching methods.
For disease and cell lines, PubTator offers access to two TaggerOne models \cite{Leaman2016} for both NER and NEN: the first trained on NCBI Disease \cite{Dogan2014} and BC5CDR corpora, the second on BioID \cite{arighi2017bio}.

\medskip

\noindent \textbf{SciSpacy} uses a pipeline approach separating the NER and NEN steps.
For NER, SciSpacy uses a neural network approach based on Stack LSTMs \cite{lample2016neural}.
It offers four distinct models trained on different biomedical corpora: BC5CDR for chemicals and diseases, CRAFT \cite{bada2012concept} for cell types, chemicals, proteins, genes, and species, JNLPBA for cell lines, cell types, DNAs, RNAs and proteins, and BioNLP13 CG \cite{pyysalo2013overview} for chemicals, diseases, genes, and species.
The normalization component included in the tool leverages a string-matching approach based on characters 3-grams. 
SciSpacy also uses a dedicated abbreviation resolution module, which identifies and expands abbreviations in the text to increase further
downstream normalization performance.
SciSpacy allows linking to multiple KBs. We use UMLS \cite{TheUnifiedMedBodenr2004} since it covers all entity types and KBs used in our cross-corpus experiments (see Section \nameref{sec:eval_framework}).
We note, however, that due to the UMLS licensing, SciSpacy comes only with a subset of the resources of UMLS, namely those categorized as in levels 0, 1, 2, and 9 from the license\footnote{\url{https://uts.nlm.nih.gov/uts/license/license-category-help.html}}.

\medskip

\noindent \textbf{HunFlair2} is an updated version of the biomedical NER tool HunFlair \cite{weber2021hunflair} which we introduce for the first time here.
Compared to the first, HunFlair2 adds support for entity normalization, allowing end-to-end entity extraction in a pipelined approach, and replaces the RNN-based character language models with transformers based on pre-trained BioLink-BERT weights \cite{YasunagaLL22}. Furthermore, it employs a single model that jointly extracts entities instead of training a separate model for each, similar to AIONER \cite{luo2023aioner}. Multi-task training uses eight corpora, i.e., BioRED \cite{luo2022biored} for annotations of all five supported entity types, NLM Gene \cite{Islamaj2021} and GNormPlus \cite{Wei2015} for genes, Linneaus \cite{gerner2010linnaeus} and S800 \cite{pafilis2013species} for species, NLM Chem \cite{islamaj2021nlm}, and SCAI Chemical \cite{kolarik2008chemical} for chemicals, and NCBI Disease \cite{Leaman2016} and SCAI disease \cite{gurulingappa2010empirical} for disease annotations. 
We indicate the entity types to extract by a predefined string sequence at the start of each input instance. 
For instance, the "\textit{[Tag genes] $<$input-example$>$}" sequence is used for extracting only gene entities, the "\textit{[Tag diseases] $<$input-example$>$}" for extracting diseases and the sequence "\textit{[Tag chemicals, diseases, genes and species] $<$input-example$>$}" is used to extract all entity types at once. 
Output labels are assigned using the standard IOB labeling scheme, with \textit{B-$<$entity type$>$} and \textit{I-$<$entity type$>$} denoting a particular type. 
A comparison of the NER performance of HunFlair2 and HunFlair can be found in Supp. Data \ref{app:hunflair_comparison}; overall, HunFlair2 improves 2.02pp over HunFlair across entity types and corpora. 
The normalization component of HunFlair2 is implemented as follows. 
We rely on the pre-trained BioSyn models released by \cite{Sung2020} similar to BERN2 for diseases, chemicals, and genes.
For species, we use SapBERT \cite{Liu2021a}, which is a BioBERT model pre-trained on UMLS. 
We note that, as in BERN2, the BioSyn gene model has been trained on the BC2GN corpus \cite{OverviewOfBioMorgan2008}, and it links exclusively to the human subset of the NCBI Gene \cite{Brown2015}, i.e., it is limited to human genes.

\section{Evaluation framework}
\label{sec:eval_framework}
This section describes the evaluation framework used to assess the performance \enquote{in the wild} of the selected entity extraction tools.
To this end, we employ a \textit{cross-corpus} evaluation protocol.
The core idea is to apply the tools to diverse texts not part of their training procedure to assess their robustness when faced with shifts in central characteristics like text types, genres, entity type definitions, or annotation guidelines.
%
%
\begin{table*}[tbph]
	\centering
 	\caption{Overview of basic information and statistics of the used evaluation corpora.
        For each data set we highlight the total number of entity mentions, the number of unique entities (Uniq.) and the knowledge base used to normalize the entities (in parenthesis).
        Note, we only highlight entity type from each data set that we leverage during our evaluation.
        $\dagger$ MedMentions originally uses UMLS for entity normalization, however, for compatibility with the evaluated tools we map the identifiers to CTD chemicals and CTD chemicals leveraging cross-reference tables and MESH identifiers.  
        }
	\resizebox{\textwidth}{!}{
		\begin{tabular}{llrr|cc|cc|cc|cc}
                    \toprule
                &
                &                    
                & 
                &
                \multicolumn{2}{c|}{\textbf{Chemical}} &
                \multicolumn{2}{c|}{\textbf{Disease}} & 
                \multicolumn{2}{c|}{\textbf{Gene}} & 
                \multicolumn{2}{c}{\textbf{Species}} 
                    \\
                      
			\textbf{Corpus} & 
                \textbf{Text type} & 
                \textbf{Documents} & 
                \textbf{Tokens} &
                \textbf{Mentions}  & \textbf{Uniq.} &
                \textbf{Mentions}  & \textbf{Uniq.} &
                \textbf{Mentions}  & \textbf{Uniq.} &
                \textbf{Mentions}  & \textbf{Uniq.}  
                        \\
			         \midrule
   
			BioID \cite{arighi2017bio} 
                & figure captions    
                & 13,697       
                & 708,913
                & \multicolumn{2}{c|}{-} 
                & \multicolumn{2}{c|}{-} 
                & \multicolumn{2}{c|}{-} 
                & 7,949 & 149 
                    \\

                & 
                &        
                & 
                & \multicolumn{2}{c|}{-} 
                & \multicolumn{2}{c|}{-} 
                & \multicolumn{2}{c|}{-} 
                & \multicolumn{2}{c}{\textit{(NCBI Taxonomy)}}
                    \\
                
                MedMentions\cite{mohanmedmentions}      
                & abstract           
                & 4,392
                & 1,012,453
                & 19,199 & 2,531
                & 19,298 & 1,694
                & \multicolumn{2}{c|}{-}            
                & \multicolumn{2}{c}{-}
                    \\

                &            
                & 
                & 
                & \multicolumn{2}{c|}{\textit{(CTD chemicals)$\dagger$}} 
                & \multicolumn{2}{c|}{\textit{(CTD diseases)$\dagger$}} 
                & \multicolumn{2}{c|}{-}            
                & \multicolumn{2}{c}{-}
                    \\
			
                tmVar (v3) \cite{Wei2022}
                & abstract
                & 500
                & 119,066
                & \multicolumn{2}{c|}{-}
                & \multicolumn{2}{c|}{-}
                & 4,059 & 677
                & \multicolumn{2}{c}{-} 
                        \\

                & 
                & 
                & 
                & \multicolumn{2}{c|}{}
                & \multicolumn{2}{c|}{}
                & \multicolumn{2}{c|}{\textit{(NCBI Gene)}}
                & \multicolumn{2}{c}{} 
                        \\

			\bottomrule
        \end{tabular}
	}
	~\label{tab:corpora:overview}
\end{table*}


\subsection{Corpora}
\label{sec:eval_framework:corpora}
Designing a cross-corpus benchmark for entity extraction poses unique challenges in terms of data selection because suitable data sets require the following conditions to be met: (a) the corpora have not been used in the training process (training or development split) of an evaluation candidate, (b) the corpora contain both NER and NEN annotations and (c) entity types are normalized to KBs supported by all tools.
Regarding (c), after reviewing the selected tools, we choose the following KBs: NCBI Gene \cite{Brown2015} for genes, CTD Diseases\footnote{CTD Diseases is also known as MEDIC and represents a subset of MeSH~\cite{lipscomb2000medical} and OMIM~\cite{hamosh2005online}} \cite{ComparativeToxDavis2023} for diseases, CTD Chemicals \cite{ComparativeToxDavis2023}\footnote{CTD chemicals represents a subset of MESH~\cite{lipscomb2000medical}.} for chemicals and NCBI Taxonomy \cite{TheNcbiTaxonoScott2012} for species.
To ensure these conditions, we select the following corpora: BioID \cite{arighi2017bio}, MedMentions \cite{mohanmedmentions}, and tmVar (v3) \cite{Wei2022}.
In MedMentions, annotated spans are linked to UMLS~\cite{TheUnifiedMedBodenr2004} instead of CTD chemicals and CTD diseases.
However, UMLS provides cross-reference tables enabling mapping their identifiers to MESH~\cite{lipscomb2000medical}.
Both CTD vocabularies represent subsets of MESH, so we map all spans in MedMentions to CTD by using their respective MESH identifiers, and we assign the entity type as either disease or chemical based on the CTD vocabulary to which the identifier has been successfully mapped.

In Table~\ref{tab:corpora:overview}, we present an overview of the corpora (see \refappendix{app:corpora} for a detailed description), which we access via BigBio \cite{fries2022bigbio}, a community library of biomedical NLP corpora. As none of the corpora is used by any of the tools, we always use the entire corpus for evaluation rather than their pre-defined test split.

\subsection{Metrics}
\label{sec:evaluation_nen}
For all tools, we report \emph{mention-level} micro (average over mentions) F1 score.
As entity extraction accounts for both recognition and normalization,
predictions and gold labels are triplets: start and end offset of the mention boundary and KB identifier.
Predicted mention boundaries are considered as correct if they differ by only one character either at the beginning or the end. 
We opt for this slightly lenient scheme to account for different handling of special characters by different tools, which may result in minor span differences (see \refappendix{app:evaluation_ner} for more details on NER evaluation).
A predicted triplet is a true positive iff both the mention boundaries and the KB identifier are correct.
Following \cite{BelbABiomediGarda2023}, for mentions with multiple normalizations, e.g., composite mentions (\enquote{breast and squamous cell neoplasms}), we consider the predicted KB identifier correct if it is equal to any of the gold ones.


\section{Results}\label{sec:results}

\begin{table*}[tbhp]
	\centering
	\caption{Mention-level cross-corpus micro F1 for named entity extraction, i.e. end-to-end entity recognition and normalization.
    Note that, SciSpacy does not support the normalization of gene, so we can't report result scores for this scenario.
		$\dagger$ Results including ChEBI annotations. 
	}
	\begin{tabular}{l|ccccc}
		\toprule
		                  & \textbf{BERN2}         & \textbf{HunFlair2} & \textbf{PubTator} & \textbf{SciSpacy} & \textbf{bent} \\
		\midrule
		\textit{Chemical} &                        &                    &                   &                   &               \\
		\quad MedMentions & 41.79 (33.42$\dagger$) & \textbf{51.17}     & 31.28             & 34.95             & 40.90         \\
		\midrule
		\textit{Disease}  &                        &                    &                   &                   &               \\
		\quad MedMentions & 47.33                  & \textbf{57.57}     & 41.11             & 40.78             & 45.94         \\
		\midrule
		\textit{Gene}     &                        &                    &                   &                   &               \\
		\quad tmVar (v3)  & 43.96                  & 76.75              & \textbf{86.02}    & -                 & 0.54          \\
		\midrule
		\textit{Species}  &                        &                    &                   &                   &               \\
		\quad BioID       & 14.35                  & 49.66              & \textbf{58.90}    & 37.14             & 10.35         \\
		\midrule
		Avg               & 36.86 (34.72$\dagger$) & \textbf{58.79}     & 54.33             & 37.61             & 24.43         \\
		\bottomrule
	\end{tabular}
	\label{tab:nen_results_ml}%
\end{table*}%

In Table \ref{tab:nen_results_ml}, we report the results (micro F1) of our cross-corpus entity extraction evaluation (see Table \ref{tab:prf1_nen} in the appendix for precision and recall).
First, we note that our results confirm previous findings \cite{giorgi2020towards,weber2020huner,weber2021hunflair}: When applied to texts different from those used during training, the performance to be expected from current tools is significantly reduced compared to published results (see Section "\nameref{sec:cross-vs-in-corpus}" for an in-depth discussion). Unlike previous studies, which considered only entity recognition, the drop in performance is even larger, which can be explained by the increased complexity of the normalization task. 
The overall best-performing model is HunFlair2, with PubTator being second, both having considerably higher average performance than the other three competitors. 
The instances where these two models stand out are the gene (tmVar v3) and species (BioID) corpora.

For genes, PubTator remarkably outperforms all models, while HunFlair2 is second.
PubTator's advantage can be explained by its highly specialized GNormPlus system usage.
In contrast, HunFlair2 here is disadvantaged because it uses only the human subset of NCBI Gene, thus being incapable of linking other species' genes.
Secondly, as HunFlair2 does not use context information for linking, 
it cannot effectively handle intra-species gene ambiguity.
For instance, depending on the context,  \enquote{TPO} can be either the 
human gene 7173 (\enquote{thyroid peroxidase}) or 7066 (\enquote{thrombopoietin}).
Though BERN2 also uses GNormPlus, its performance is notably lower than PubTator's.
%
This is attributable to its NER component, which introduces many false positives (see Discussion).
SciSpacey achieves a sub-par performance for chemicals and diseases while scoring third best for species. 
Finally, we note bent's exceptionally low score on genes.
By inspecting its predictions, we find that this is due to the tool consistently predicting genes of species that are not human.
For instance, all mentions of \enquote{BRCA1}, instead of being linked to the human gene (672), are linked to the Capuchin monkey (108289781).
As 96\% of mentions in tmVar (v3) are human genes, this drastically impacts bent's results.

Regarding species, the leading cause for the low performance of BERN and bent are subtle differences in the KB identifiers, primarily for mentions of the mouse concept.
Mouse is one of the organisms most frequently mentioned in biomedical publications (see Discussion). In BioID, its mentions are linked to NCBI Taxonomy 10090 \enquote{house mouse}.
While both PubTator and HunFlair2 also return 10090, bent links mentions of mouse to NCBI Taxonomy 10088 (\enquote{mouse genus}), while BERN2 to 10095 (\enquote{unspecified mouse}), causing a drastic drop in performance.

For diseases, we see that differences are not as pronounced
with almost all tools achieving $>$40\% F1 score, where we attribute HunFlair2's advantage to its superior NER performance (see below).
Interestingly, BERN2 comes as a close second. We hypothesize this is due to the better performance of its neural normalization components for diseases and chemicals.

\section{Discussion}\label{sec:discussion}

\subsection{Cross-corpus vs. in-corpus evaluation}
\label{sec:cross-vs-in-corpus}
Evaluating NER and NEN together is of high importance for applications of BTM, as mentions without assigned ID cannot be easily integrated with other information, like entity annotations, experimental results, or findings of the same entity in other documents. Accordingly, any downstream application requires a normalized entity as input, mostly achieved using complicated, tailor-made, and difficult-to-maintain combinations of specialized NER and NEN tools. Only recently have tools become available to perform both steps stream-lined and easy to use. However, their application performance remains unknown as published results are usually obtained using an in-corpus evaluation. Here, five such tools were compared in a cross-corpus protocol for the first time.  

In an in-corpus evaluation, training and test data come from the same source, ensuring the model is evaluated on instances closely resembling those seen during training, ensuring consistency in terms of concept definitions, annotation guidelines, and scientific scope and purpose of the analysis. This setting is beneficial when the goal is to measure the model's performance on the same domain as the training data, i.e., when the training data was specifically created for the downstream application.

However, consistency between training data and application data cannot be assumed in a setting where readily trained models are applied to new texts, i.e., in real-life applications of BTM tools. In-corpus evaluations cannot accurately represent the model's ability to generalize to such settings and will most probably underestimate the model's limitations in handling variations. Cross-corpus evaluations are a step towards obtaining more reliable estimates on the performance one can expect in real-world applications, where trained models are applied to new text collections varying in focus, entity distribution, genre or text type from the training set \cite{giorgi2020towards, tikk2010comprehensive}. 

\begin{table}[t!]
	\centering
	\caption{We show the differences between in- and cross-corpus performances of biomedical entity extraction tools on the basis of the results reported by PubTator~\cite{weiPubTatorCentralAutomated2019} and BERN2~\cite{Bern2AnAdvanMujeen}.
		Note that the in-corpus results examine the NEN components exclusively and not (as in our study) the end-to-end task of NER and NEN combined.
		For each tool and entity type we also provide the used evaluation corpus in parenthesis.
		Since the tools leverage different corpora for evaluation the results are not directly comparable, but they indicate the magnitude of the in-corpus, cross-corpus performance deviations.
		$\dagger$ Result represents accuracy scores.
	}
	\begin{tabular}{l|ll}
		\toprule
		\textbf{Tool}           & \textbf{In-corpus}                & \textbf{Cross-corpus}        \\
		\midrule

		BERN2                   &                                   &                              \\
		\quad \textit{Chemical} & 96.60$\dagger$ \textit{(BC5CDR)}  & 41.68 \textit{(MedMentions)} \\
		\quad \textit{Disease } & 93.90$\dagger$  \textit{(BC5CDR)} & 47.31 \textit{(MedMentions)} \\
		\quad \textit{Gene    } & 95.90$\dagger$ \textit{(BC2GM)}   & 43.81 \textit{(tmVar v3)}    \\
		\midrule
		PubTator                &                                   &                              \\
		\quad \textit{Chemical} & 77.20 \textit{(NLM-Chem)}         & 31.26 \textit{(MedMentions)} \\
		\quad \textit{Disease } & 80.70 \textit{(NCBI-Disease)}     & 40.76 \textit{(MedMentions)} \\
		\quad \textit{Gene    } & 72.70 \textit{(NLM-Gene)}         & 85.92 \textit{(tmVar v3)}    \\
		\bottomrule
	\end{tabular}
	\label{tab:in_corpus_results}%
\end{table}%

However, the question remains how large the difference between in-corpus and cross-corpus evaluations actually is. To approach an answer to this question for entity extraction, Table \ref{tab:in_corpus_results} compares published in-corpus and our newly measured cross-corpus results for BERN2 and PubTator for the three entity types genes, chemicals, and species. Both tools generally report drastically higher scores for in-corpus performance compared to our results in a cross-corpus regime. Compared to cross-corpus evaluations only for NER \cite{giorgi2020towards, weber2020huner, weber2021hunflair}, differences mostly grew further, often by a large margin. Note that this comparison omitted the catastrophic yet easy-to-explain results reported for the same cases in Table~\ref{tab:nen_results_ml}. A notable exception, which again points to the difficulties of such comparisons, is the extraction of genes using PubTator. Here, we report a higher score (85.92) for our cross-corpus than in the published in-corpus results (72.70). This difference can be attributed to the different
species distribution of the mentioned genes. In the tmVar corpus (used in our study), $\sim$96\% of the mentioned genes correspond
to humans. In contrast, NLM-Gene shows a more diverse distribution consisting of only 48\% human genes, thus being a significantly more complex test case due to cross-species gene ambiguity \cite{CrossSpeciesGWeiC2011}.

A cross-corpus setting, however, is also subject to several limitations, as it measures performance on yet only one other corpus. Results will vary depending on the corpus used and might still be far off what one can expect in real life. These differences are particularly relevant in the field of biomedical NLP as there are no standardized concept definitions or shared annotation guidelines (e.g., some corpora distinguish between proteins and genes \cite{collier2004introduction} while others do not \cite{islamaj2021nlm}). This lack of standardization leads to inconsistencies and discrepancies when comparing model performance across corpora, making it difficult to draw reliable conclusions about the model's generalization capabilities. In this sense, our study can only show that the expected differences are large, but we cannot reliably estimate the actual performance one can expect for a specific application. The only reliable yet laborious way of obtaining such estimates is the usage of an application-specific annotated corpus.


\subsection{Open issues for fair comparions}

Establishing a robust and fair evaluation framework for biomedical NER and NEN presents unique challenges. This section discusses some of them and explains our choices for choosing an option when multiple ones exist. Clearly, opting for another option for comparisons would lead to different results, as would choosing other evaluation corpora.

One major issue is annotation consistency between multiple corpora. Single corpus annotations may already suffer from inter-annotator disagreement, i.e., multiple annotators carry out annotations on the same corpus and mark different entities as correct. Usually, multiple rounds of annotations are introduced to reduce this problem and to resolve ambiguities \cite{Islamaj2021}.
However, annotation consistency across multiple corpora is a much more challenging topic as annotation guidelines differ, and the scopes of the individual corpora vary. This often leads to conflicting annotations, for instance, when one entity must be annotated under one guideline and not under another. Even if an entity should be annotated under two different guidelines, the exact entity boundaries may differ depending on the entity definitions regarding entity attributes and adjectives. This is particularly apparent in a cross-corpus setting, as entity boundaries on which the tools are trained may deviate significantly from those in the evaluation corpora. We chose not to analyze differences in annotation guidelines further prior to our study as this probably would have made any comparison impossible, given the scarcity of corpora that were not used by any of the considered candidate tools during training. 

The handling of non-consecutive and/or overlapping spans is closely related to this aspect. For instance, for the text segment ``\textit{[...] is causing breast and ovarian cancer [...]}'', there is no universally agreed upon guideline on how to annotate the two disease mentions. 
Different corpora use different strategies, e.g.:
\begin{itemize}
	\item ``\textit{breast}'' and ``\textit{ovarian cancer}'' as two distinct entities,
	\item ``\textit{breast and ovarian cancer}'' as one entity having two normalization identifiers assigned,
	\item ``\textit{breast cancer}'' and ``\textit{ovarian cancer}'', where the former is an entity consisting of
	      multiple, non-consecutive text segments.
\end{itemize}
Opting for one strategy vs another inevitably introduces a bias towards or against a specific tool.
Our study follows the common practice of excluding non-consecutive entities from evaluation. 

A third choice for evaluation is whether one wants to consider spans at all. Note that for many applications, such as information retrieval~\cite{perera2020named} or semantic indexing 
 \cite{ChemicalIdentiLeaman2023}, mention boundaries are not relevant because \textit{document-level} predictions suffice. We also performed an additional evaluation for capturing such a setting, whose results are shown in Figure \ref{fig:ablation_nen}. All tools achieve notable performance gains (up to 4 pp), which can be explained by the different and simpler task, as models are not required anymore to identify (a) every entity mentioned and (b) find mention boundaries (see \refappendix{app:nen_doc_level} for details).

A fourth issue is related to the KBs used as targets for the normalization step. Different corpora and tools often use different KBs, so a mapping of identifiers is necessary before a comparison. However, as different KBs can implement different definitions of entity types, these mappings may lack completeness and precision \cite{Vasilevsky2020}, which impacts performance metrics. Secondly, such cross-reference tables are not always available and are challenging to build according to diverging naming conventions, hierarchies, and syntactical rules. In our case, the choice of KB was straightforward for genes, diseases, and species, as tools and corpora essentially coincided with these types. However, we had to take more difficult decisions for chemicals, because BERN2 normalizes chemicals to CTD Chemicals (65\%) and ChEBI (35\%). However, to the best of our knowledge, no mapping is available between ChEBI and CTD. This required us to modify the BERN2 installation to only use CTD, ignoring many of its normalization results. 

A fifth issue is that minimal variations in the normalization choices can introduce substantial differences, as exemplified by the \enquote{mouse} case reported in \nameref{sec:results}. 
The issue can be mitigated by the use of evaluation metrics taking into account the KB hierarchy. 
For example, the measures introduced by \cite{EvaluationMeasKosmop2015} consider the lowest common ancestor for the evaluation. 
It considers a limited set of ancestors and penalizes predictions according to the hierarchical classification scheme. 
This is, however, limited to those KBs structured hierarchically, which is not always the case (e.g., NCBI Gene).

Finally, we note that technical issues related to the tool's implementation may hinder straightforward comparisons. 
For instance, PubTator and BERN2 had issues processing non-ascii special characters, which required extensive preprocessing. 

\subsection{Performance of NER step}\label{results:ner}

\begin{figure*}[!tbhp]
    \centering
    \subfloat[][]{
        \includegraphics[width=0.49\textwidth]{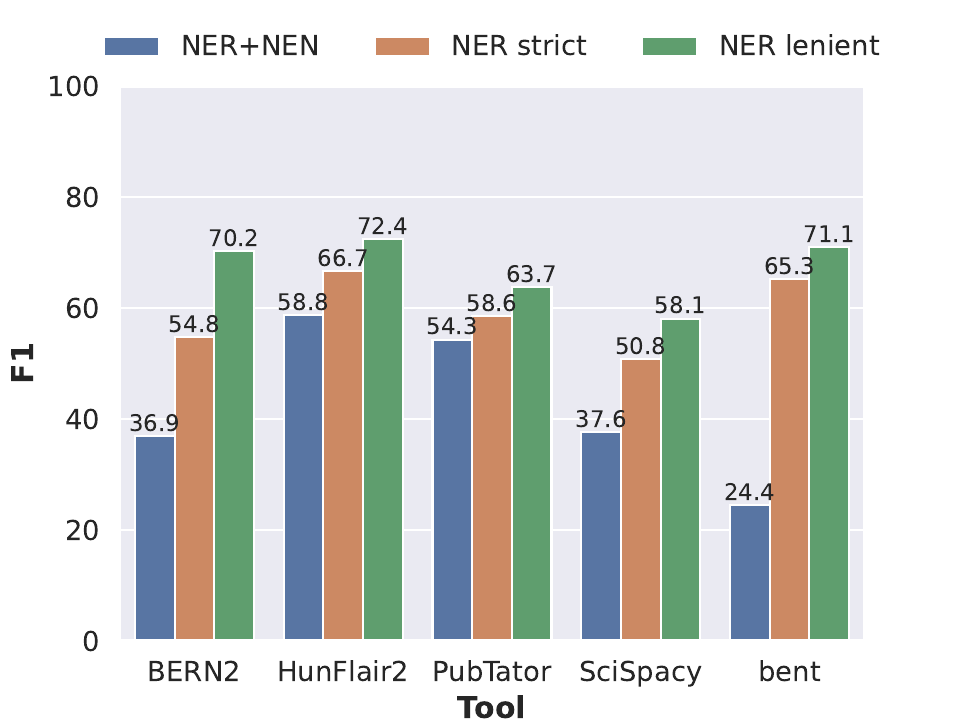}
        \label{fig:ablation_ner}}
    \subfloat[][]{
        \includegraphics[width=0.49\textwidth]{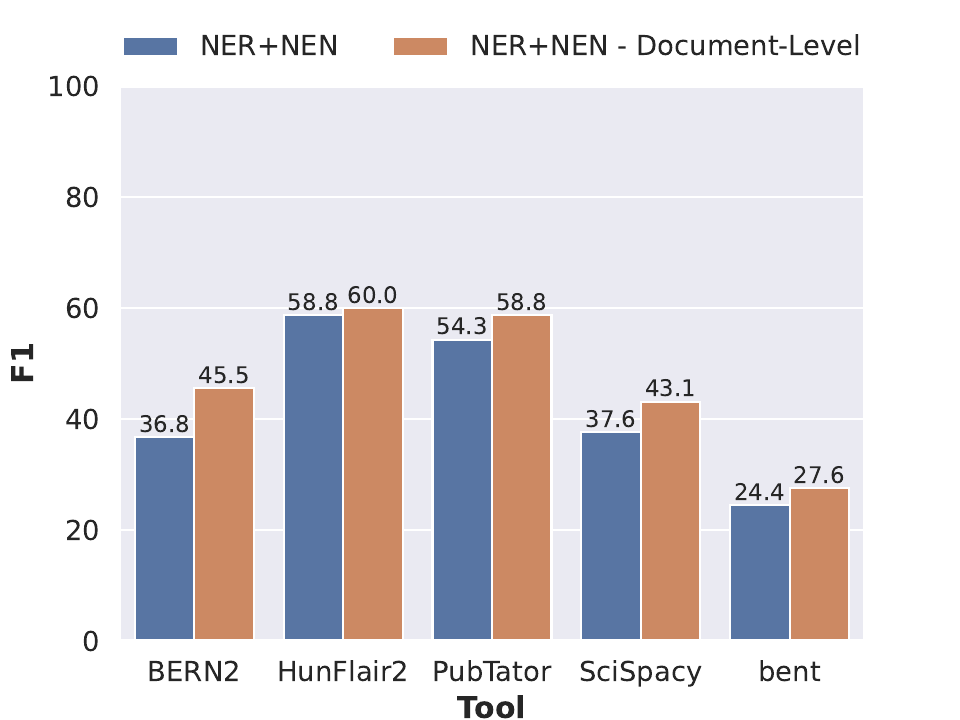}
        \label{fig:ablation_nen}}
    \caption{Ablation study results: (a) Performance comparison of the five tools concerning three evaluation settings: end-to-end NER and NEN, NER using a strict and a lenient evaluation setting, i.e., we count each prediction as true positive which is a sub- or superstring of gold standard entity mention. (b) Comparison of the mention- and document-level end-to-end NER and NEN results of the five tools.  }
    \label{fig:ablation_results}
\end{figure*}

As the differences we observed between measured and published performance results across all tools exceed those reported previously for cross-corpus evaluations of NER, we aimed to quantify the errors introduced exclusively by wrong normalizations more precisely. To this end, we compared the NER+NEN results to those of two "NER only" evaluations, where we once required, as before, entity boundaries to mismatch by at most one character (strict), and once only required that the predicted span is completely contained in the annotated span or vice versa (lenient). Results are shown in Figure \ref{fig:ablation_ner}.

Regarding our strict setting, HunFlair2 retains its leading position in the average score across all entity types. However, Pubtator is now superseded by bent regarding pure NER performance. Inspecting results by corpus (see \refappendix{app:ner_results}) reveals that in this setting, PubTator's NER performance is severely reduced by its comparably low performance for chemicals. The detection quality of gene mentions varies widely, with a range from 48.05\% (BERN2) to
88.82\% (PubTator). SciSpacy, in this evaluation, proves to be a valid tool, though still inferior to the other four candidates, showing that its low performance in the extraction setting is primarily rooted in its sub-par performance in entity normalization. 

In the lenient setting, the performance for all tools increases further, on average by 7 pp. Hunflair still heads the performance, very closely followed by bent and BERN2, which overtakes Pubtator in this setting.
The strongest performance improvement is recorded for BERN2, whose F1 increases by 15.4 pp from 54.8 percent to 70.2 percent. 
This gain can be mainly attributed to the identification of genes for which
the F1 score increases by 40.7 pp (see \refappendix{app:ner_results}). 
A closer inspection of the BERN2 predictions reveals that the quality improvements are mainly due to its handling of the token "gene". When the term \enquote{gene} immediately succeeds a gene mention (e.g., \enquote{AKT-1 gene}), BERN2 often incorporates the term in its prediction, whereas the gold standard corpus does not.

\subsection{Infrequent entities}\label{sec:entity_distribution}
\begin{table*}[tbhp]
	\caption{Distribution statistics of the top five disease entities in MedMentions (left) and species entities in BioID (right) with the corresponding three most frequent mentions.}
	\label{tab:entity_distribution}
	\begin{minipage}{\columnwidth}
		\begin{tabular}{llll}
			\toprule
			\textbf{Entity} & \textbf{Count (\%)} & \textbf{Mention} & \textbf{Count (\%)} \\
			\midrule
			MESH:D009369    & 715 (3.89\%)        & cancer           & 177 (0.96\%)        \\
			                &                     & tumor            & 169 (0.92\%)        \\
			                &                     & tumors           & 105 (0.57\%)        \\
			\midrule
			MESH:D004194    & 537 (2.92\%)        & disease          & 238 (1.3\%)         \\
			                &                     & diseases         & 103 (0.56\%)        \\
			                &                     & disorders        & 46 (0.25\%)         \\
			\midrule
			MESH:D007239    & 305 (1.66\%)        & infection        & 179 (0.97\%)        \\
			                &                     & infections       & 57 (0.31\%)         \\
			                &                     & infected         & 12 (0.07\%)         \\
			\midrule
			MESH:D009765    & 262 (1.43\%)        & obesity          & 162 (0.88\%)        \\
			                &                     & obese            & 60 (0.33\%)         \\
			                &                     & Obesity          & 25 (0.14\%)         \\
			\midrule
			MESH:D001943    & 251 (1.37\%)        & breast cancer    & 178 (0.97\%)        \\
			                &                     & BC               & 28 (0.15\%)         \\
			                &                     & Breast Cancer    & 10 (0.05\%)         \\
			\bottomrule
		\end{tabular}
	\end{minipage}%
	\hspace{0.5em}
	\begin{minipage}{\columnwidth}
		\begin{tabular}{llll}
			\toprule
			\textbf{Entity}  & \textbf{Count (\%)} & \textbf{Mention} & \textbf{Count (\%)} \\
			\midrule
			NCBI taxon:10090 & 4002 (50.35\%)      & mice             & 2923 (36.77\%)      \\
			                 &                     & mouse            & 396 (4.98\%)        \\
			                 &                     & Mice             & 64 (0.81\%)         \\
			\midrule
			NCBI taxon:9606  & 688 (8.66\%)        & human            & 228 (2.87\%)        \\
			                 &                     & patients         & 163 (2.05\%)        \\
			                 &                     & patient          & 119 (1.5\%)         \\
			\midrule
			NCBI taxon:7227  & 298 (3.75\%)        & flies            & 134 (1.69\%)        \\
			                 &                     & larvae           & 53 (0.67\%)         \\
			                 &                     & fly              & 18 (0.23\%)         \\
			\midrule
			NCBI taxon:4932  & 196 (2.47\%)        & S. cerevisiae    & 48 (0.6\%)          \\
			                 &                     & yeast            & 40 (0.5\%)          \\
			                 &                     & Yeast            & 17 (0.21\%)         \\
			\midrule
			NCBI taxon:7955  & 179 (2.25\%)        & larvae           & 72 (0.91\%)         \\
			                 &                     & embryos          & 48 (0.6\%)          \\
			                 &                     & zebrafish        & 34 (0.43\%)         \\
			\bottomrule
		\end{tabular}
	\end{minipage}%
\end{table*}

Corpora are usually designed for specific sub-domains or applications, e.g., the PDR corpus \cite{cho2017method} focuses on plants, whereas BioNLP2013-CG \cite{pyysalo2013overview} is concerned with cancer genetics.
Additionally, the annotated documents may reflect the predominance of specific topics or research trends, e.g., oncology in case of diseases \cite{RegelCorpusIGarda2022}, in the literature. 
Consequently, their documents often present imbalanced entity distributions, with few highly frequent entities and a long tail of rarely occurring ones.
To illustrate this more concretely, Table \ref{tab:entity_distribution} shows the five most frequent species and disease entities (with their three most frequent mentions) in BioID and MedMentions, respectively. 
The most frequent species entity in BioID is NCBI Taxonomy 10090, accounting for more than half of all species mentions in the corpus. 
Similarly, MedMentions has a bias towards cancer (MESH:D009369), accounting for $\sim$4\% of all disease mentions. 
These imbalances raise the question of how much the performance can be attributed to correctly extracting the most frequently occurring entities.


\begin{table*}[tbhp]
	\centering
	\caption{Macro F1 scores for named entity extraction, i.e., end-to-end entity recognition and normalization.
		We compute precision and recall for each entity in the KB and take the harmonic mean (F1).
		We compare the results to micro F1 scores in Table \ref{tab:nen_results_ml} (difference in brackets).}
	\label{tab:nen_results_entity_distribution}
		\begin{tabular}{lrrrrr}
		\toprule
		                  & \textbf{BERN2}    & \textbf{HunFlair2} & \textbf{PubTator} & \textbf{SciSpacy} & \textbf{bent}     \\
		\midrule
		\textit{Chemical} &                   &                    &                   &                   &                   \\
		\quad MedMentions & 30.15             & \textbf{33.27}              & 21.93             & 32.13             & 27.86             \\
		                  & (\textit{-11.64}) & (\textit{-17.90})  & (\textit{-9.35 }) & (\textit{-2.82 }) & (\textit{-13.04}) \\
		\midrule
		\textit{Disease}  &                   &                    &                   &                   &                   \\
		\quad MedMentions & 30.43             & \textbf{38.47}              & 26.14             & 36.65             & 23.96             \\
		                  & (\textit{-16.90}) & (\textit{-19.10})  & (\textit{-14.97}) & (\textit{-4.13 }) & (\textit{-21.98}) \\
		\midrule
		\textit{Gene}     &                   &                    &                   &                   &                   \\
		\quad tmVar (v3)  & 25.51             & 47.47              & \textbf{72.46}             & -             & 0.31              \\
		                  & (\textit{-18.45}) & (\textit{-29.28})  & (\textit{-13.56}) & (\textit{-}) & (\textit{-0.23 }) \\
		\midrule
		\textit{Species}  &                   &                    &                   &                   &                   \\
		\quad BioID       & 16.83             & 8.38               & \textbf{20.73}             & 4.78              & 3.25              \\
		                  & (\textit{+2.48 }) & (\textit{-41.28})  & (\textit{-38.17}) & (\textit{-32.36}) & (\textit{-7.10 }) \\
		\midrule
		Avg               & 25.73             & 31.90              & \textbf{35.32}             & 24.52             & 13.84             \\
		                  & (\textit{-11.13}) & (\textit{-26.89})  & (\textit{-19.01}) & (\textit{-13.10 }) & (\textit{-10.59}) \\
		\bottomrule
	\end{tabular}
\end{table*}

To answer this, we compute macro F1 scores for each entity type by calculating individual scores for each unique entity and then averaging them. As an example for the species entity \enquote{NCBI Taxon 10090} (mouse), we calculate distinct precision and recall scores from all occurrences in different text spans: Two correctly predicted mentions of \enquote{mouse} (true positives)
and \enquote{mice}
with no further false positive and false negative predictions, this entity would have an F1 score of 1. For detailed information about macro F1-score computation, we refer to the \refappendix{app:macro_f1_score}.
From Table \ref{tab:nen_results_entity_distribution}, we observe a notable drop in F1 from all tools across all entity types, highlighting how an important fraction of the overall performance can indeed be explained by the tools correctly identifying frequently occurring entities. The tool performing best in this setting is PubTator with HunFlair2 reaching the second position, which can probably be explained by the dictionary-based components in PubTators normalization modules that can better deal with rare entity names, especially those not seen during training. 
Most notable is the strong performance degradation for species, which is also consistent with the entity distribution information from Table~\ref{tab:entity_distribution}. 
We refer to \refappendix{app:infrequent_entities_ner} for a similar analysis on NER results.

\subsection{Overlap analysis across tools}
To identify shared features and differences of the tools evaluated, we examine the overlap of their true positive extracted entities. Results are shown per entity type in Figure~\ref{fig:tp_overlaps} in the Appendix.
We observe a relatively high agreement in normalizing genes and species, showing an overlap of 40.3\% and 43.5\% of entities correctly found by all tools.
The agreement for chemical and disease mentions is slightly lower (34.6\% / 34.3\%), reflecting the more pronounced performance differences shown in Table~\ref{tab:nen_results_ml}.
When looking at the correct predictions per entity type exclusively detected by only one tool, the ratio is notably high for chemicals (17.9\%) and species (18.1\%), moderate for diseases (8.3\%), and low for gene mentions (4.2\%).
The high value for chemicals and species can be attributed to individual tools, HunFlair2 (11.8\% for chemicals) and bent (11.1\% for species).
We inspected these exclusive true positive predictions for both tools, highlighting 513 / 324 unique chemical mentions/identifiers for HunFlair2 and 69 / 44 unique species mentions and identifiers for bent.
The most frequent exclusive chemical predictions by HunFlair concern \textit{lipids} (MESH:D008055, 98~mentions), \textit{reactive oxygen species} (MESH:D017382, 87), \textit{lipopolysaccharides} (MESH:D008070, 54), and \textit{water} (MESH:D014867, 47).
In case of species mentions detected by bent only, \textit{mus musculus} (NCBI~taxon:10090, 183), \textit{homo sapiens} (NCBI~taxon:9606, 114), \textit{ hepacivirus hominis} (NCBI~taxon:11103, 39), and \textit{adenovirus} (NCBI~taxon:10535, 36) represent the most frequent entities.
These results indicate that, especially for chemicals and species, the tools reveal a heterogeneous prediction profile and that ensembling the methods could improve recall.

\section{Conclusion}
\label{sec:conclusion}
In this work, we reviewed 28 recent tools designed for extracting biomedical named entities from unstructured text regarding their maturity and ease of usage for downstream applications. We selected five tools, namely BERN2, bent, HunFlair2, PubTator, and SciSpacy, for a detailed examination and assessed their performance on three different corpora, encompassing four types of entities, following a cross-corpus approach. 
Our experiments highlight that the performance of the tools varies considerably across corpora and entity types. 
Additionally, we found strong performance drops compared to the published in-corpus results.
In-depth prediction analyses revealed that the tools demonstrate strong performance when identifying highly researched entities; however, they face challenges in accurately identifying concepts that rarely occur in the literature. 
In conclusion, our results illustrate that further research is needed on the generalization capabilities of named entity extraction tools to facilitate their seamless application to diverse biomedical subdomains and text types. \\

\medskip

\noindent\fbox{\parbox{0.94\linewidth}{%
\textbf{Key points:}
\begin{itemize}
    \item In-depth cross-corpus evaluation of five recent deep learning-based entity extraction tools
    \item Update of the HunFlair tool to support entity recognition and normalization using transformer-based models
    \item Experimental results highlight considerable performance variations compared to published in-corpus results
    \item All five tools encounter difficulties in accurately identifying concepts rarely occurring in the literature
\end{itemize}
}}
%

\section{Competing interests}
No competing interest is declared.

\section{Author contributions statement}
M.S., S.G., P.D., B.F., and A.A. implemented entity normalization in HunFlair2. 
X.W. and L.W-G. re-trained and extended the HunFlair2 named entity recognition models.
M.S., S.G., and X.W. conceived the experiments, analyzed the results and wrote the (initial version of the) manuscript.
U.L., L.W-G., P.D., B.F., and A.A. reviewed and revised the manuscript.
M.S., L.W-G. and U.L. conceived the project.
U.L. supervised and administrated the project and acquired financial support.


\section{Acknowledgments}
Mario Sänger is supported by Deutsche Forschungsgemeinschaft (DFG, German
Research Foundation) CRC 1404: "FONDA: Foundations of Workflows for
Large-Scale Scientific Data Analysis". 
Samuele Garda is supported by the DFG as part of the research unit \enquote{Beyond the Exome}. 
Xing David Wang is supported by the DFG as part of the research unit CompCancer (No. RTG2424).
Alan Akbik is supported by the DFG under Germany's Excellence Strategy "Science of Intelligence" (EXC 2002/1,
project number 390523135) and the DFG Emmy Noether grant "Eidetic Representations of Natural Language" (project number 448414230).



\bibliographystyle{plain}
\bibliography{reference}

\begin{thebibliography}{100}

\bibitem{Agarwal2022}
Dhruv Agarwal, Rico Angell, Nicholas Monath, and Andrew McCallum.
\newblock Entity linking via explicit mention-mention coreference modeling.
\newblock In {\em Proceedings of the 2022 Conference of the North American
  Chapter of the Association for Computational Linguistics: Human Language
  Technologies}, page 4644\textendash{}4658. Association for Computational
  Linguistics, 2022.

\bibitem{ahmed2023easyner}
Rafsan Ahmed, Petter Berntsson, Alexander Skafte, Salma~Kazemi Rashed, Marcus
  Klang, Adam Barvesten, Ola Olde, William Lindholm, Antton~Lamarca
  Arrizabalaga, Pierre Nugues, et~al.
\newblock Easyner: A customizable easy-to-use pipeline for deep learning-and
  dictionary-based named entity recognition from medical text.
\newblock {\em arXiv preprint arXiv:2304.07805}, 2023.

\bibitem{AkhondiHHMK15}
Saber~A. Akhondi, Kristina~M. Hettne, Eelke van~der Horst, Erik~M. van
  Mulligen, and Jan~A. Kors.
\newblock Recognition of chemical entities: combining dictionary-based and
  grammar-based approaches.
\newblock {\em J. Cheminformatics}, 7({S-1}):S10, 2015.

\bibitem{alshaikhdeeb2016biomedical}
Basel Alshaikhdeeb and Kamsuriah Ahmad.
\newblock Biomedical named entity recognition: a review.
\newblock {\em International Journal on Advanced Science, Engineering and
  Information Technology}, 6(6):889--895, 2016.

\bibitem{Angell2021}
Rico Angell, Nicholas Monath, Sunil Mohan, Nishant Yadav, and Andrew McCallum.
\newblock Clustering-based inference for biomedical entity linking.
\newblock In {\em Proceedings of the 2021 Conference of the North American
  Chapter of the Association for Computational Linguistics: Human Language
  Technologies}, page 2598\textendash{}2608. Association for Computational
  Linguistics, 2021.

\bibitem{arighi2017bio}
Cecilia Arighi, Lynette Hirschman, Thomas Lemberger, Samuel Bayer, Robin
  Liechti, Donald Comeau, and Cathy Wu.
\newblock Bio-id track overview.
\newblock In {\em BioCreative VI Challenge Evaluation Workshop}, volume 482,
  page 376, 2017.

\bibitem{aronson2010overview}
Alan~R Aronson and Fran{\c{c}}ois-Michel Lang.
\newblock An overview of metamap: historical perspective and recent advances.
\newblock {\em Journal of the American Medical Informatics Association},
  17(3):229--236, 2010.

\bibitem{bada2012concept}
Michael Bada, Miriam Eckert, Donald Evans, Kristin Garcia, Krista Shipley,
  Dmitry Sitnikov, William~A Baumgartner, K~Bretonnel Cohen, Karin Verspoor,
  Judith~A Blake, et~al.
\newblock Concept annotation in the craft corpus.
\newblock {\em BMC bioinformatics}, 13(1):1--20, 2012.

\bibitem{badenes2022overview}
Carlos Badenes-Olmedo, {\'A}lvaro Alonso, and Oscar Corcho.
\newblock An overview of drugs, diseases, genes and proteins in the cord-19
  corpus.
\newblock {\em Procesamiento del Lenguaje Natural}, 69:165--176, 2022.

\bibitem{Onthefly20ABaltou2021}
Fotis~A Baltoumas, Sofia Zafeiropoulou, Evangelos Karatzas, Savvas Paragkamian,
  Foteini Thanati, Ioannis Iliopoulos, Aristides~G Eliopoulos, Reinhard
  Schneider, Lars~Juhl Jensen, Evangelos Pafilis, and Georgios~A Pavlopoulos.
\newblock Onthefly2.0: a text-mining web application for automated biomedical
  entity recognition, document annotation, network and functional enrichment
  analysis.
\newblock {\em NAR Genomics and Bioinformatics}, 3, 10 2021.

\bibitem{Bhowmik2021}
Rajarshi Bhowmik, Karl Stratos, and Gerard de~Melo.
\newblock Fast and effective biomedical entity linking using a dual encoder.
\newblock In {\em Proceedings of the 12th International Workshop on Health Text
  Mining and Information Analysis}, page 28\textendash{}37. Association for
  Computational Linguistics, 2021.

\bibitem{TheUnifiedMedBodenr2004}
O.~Bodenreider.
\newblock The unified medical language system (umls): integrating biomedical
  terminology.
\newblock {\em Nucleic Acids Research}, 32:267D--270, 1 2004.

\bibitem{Brown2015}
Garth~R. Brown, Vichet Hem, Kenneth~S. Katz, Michael Ovetsky, Craig Wallin,
  Olga Ermolaeva, Igor Tolstoy, Tatiana Tatusova, Kim~D. Pruitt, Donna~R.
  Maglott, and Terence~D. Murphy.
\newblock Gene: a gene-centered information resource at ncbi.
\newblock {\em Nucleic Acids Research}, 43(D1):D36\textendash{}D42, 2015.

\bibitem{AModularFrameCampos2013}
David Campos, S\'{e}rgio Matos, and Jos\'{e}~Lu\'{\i}s Oliveira.
\newblock A modular framework for biomedical concept recognition.
\newblock {\em BMC Bioinformatics}, 14, 12 2013.

\bibitem{campos_gimli_2013}
David Campos, Sérgio Matos, and José~Luís Oliveira.
\newblock Gimli: open source and high-performance biomedical name recognition.
\newblock {\em {BMC} Bioinformatics}, 14(1):54, 2013.

\bibitem{Chen2021}
Lihu Chen, Ga\"{e}l Varoquaux, and Fabian~M. Suchanek.
\newblock A lightweight neural model for biomedical entity linking.
\newblock {\em Proceedings of the AAAI Conference on Artificial Intelligence},
  35(14):12657\textendash{}12665, 2021.

\bibitem{cho2017method}
Hyejin Cho, Wonjun Choi, and Hyunju Lee.
\newblock A method for named entity normalization in biomedical articles:
  application to diseases and plants.
\newblock {\em BMC bioinformatics}, 18(1):1--12, 2017.

\bibitem{ChoHPP20}
Minsoo Cho, Jihwan Ha, Chihyun Park, and Sanghyun Park.
\newblock Combinatorial feature embedding based on {CNN} and {LSTM} for
  biomedical named entity recognition.
\newblock {\em J. Biomed. Informatics}, 103:103381, 2020.

\bibitem{cohen2017colorado}
K~Bretonnel Cohen, Karin Verspoor, Kar{\"e}n Fort, Christopher Funk, Michael
  Bada, Martha Palmer, and Lawrence~E Hunter.
\newblock The colorado richly annotated full text (craft) corpus: Multi-model
  annotation in the biomedical domain.
\newblock {\em Handbook of Linguistic annotation}, pages 1379--1394, 2017.

\bibitem{collier2004introduction}
Nigel Collier and Jin-Dong Kim.
\newblock Introduction to the bio-entity recognition task at jnlpba.
\newblock In {\em Proceedings of the International Joint Workshop on Natural
  Language Processing in Biomedicine and its Applications (NLPBA/BioNLP)},
  pages 73--78, 2004.

\bibitem{DangLNV18}
Thanh~Hai Dang, Hoang{-}Quynh Le, Trang~M. Nguyen, and Sinh~T. Vu.
\newblock {D3NER:} biomedical named entity recognition using crf-bilstm
  improved with fine-tuned embeddings of various linguistic information.
\newblock {\em Bioinform.}, 34(20):3539--3546, 2018.

\bibitem{ComparativeToxDavis2023}
Allan~Peter Davis, Thomas~C Wiegers, Robin~J Johnson, Daniela Sciaky, Jolene
  Wiegers, and Carolyn~J Mattingly.
\newblock Comparative toxicogenomics database (ctd): update 2023.
\newblock {\em Nucleic Acids Research}, 51:D1257--D1262, 1 2023.

\bibitem{demner-fushman_metamap_2017}
Dina Demner-Fushman, Willie~J Rogers, and Alan~R Aronson.
\newblock {MetaMap} lite: an evaluation of a new java implementation of
  {MetaMap}.
\newblock {\em Journal of the American Medical Informatics Association},
  24(4):841--844, 2017.

\bibitem{SieveBasedEntDsouz2015}
Jennifer D\ensuremath{'}Souza and Vincent Ng.
\newblock Sieve-based entity linking for the biomedical domain.
\newblock In {\em Proceedings of the 53rd Annual Meeting of the Association for
  Computational Linguistics and the 7th International Joint Conference on
  Natural Language Processing (Volume 2: Short Papers)}, pages 297--302,
  Beijing, China, July 2015. Association for Computational Linguistics.

\bibitem{dernoncourt2017neuroner}
Franck Dernoncourt, Ji~Young Lee, and Peter Szolovits.
\newblock Neuroner: an easy-to-use program for named-entity recognition based
  on neural networks.
\newblock In {\em Proceedings of the 2017 Conference on Empirical Methods in
  Natural Language Processing: System Demonstrations}, pages 97--102, 2017.

\bibitem{Dogan2014}
Rezarta~Islamaj Do\u{g}an, Robert Leaman, and Zhiyong Lu.
\newblock Ncbi disease corpus: A resource for disease name recognition and
  concept normalization.
\newblock {\em Journal of Biomedical Informatics}, 47:1\textendash{}10,
  2014-02.

\bibitem{eyre2021launching}
Hannah Eyre, Alec~B Chapman, Kelly~S Peterson, Jianlin Shi, Patrick~R Alba,
  Makoto~M Jones, Tamara~L Box, Scott~L DuVall, and Olga~V Patterson.
\newblock Launching into clinical space with medspacy: a new clinical text
  processing toolkit in python.
\newblock In {\em AMIA Annual Symposium Proceedings}, volume 2021, page 438.
  American Medical Informatics Association, 2021.

\bibitem{AnOverviewOfFrench2022}
Evan French and Bridget~T. McInnes.
\newblock An overview of biomedical entity linking throughout the years.
\newblock {\em Journal of Biomedical Informatics}, page 104252, 12 2022.

\bibitem{fries2022bigbio}
Jason Fries, Leon Weber, Natasha Seelam, Gabriel Altay, Debajyoti Datta,
  Samuele Garda, Sunny Kang, Rosaline Su, Wojciech Kusa, Samuel Cahyawijaya,
  et~al.
\newblock Bigbio: a framework for data-centric biomedical natural language
  processing.
\newblock {\em Advances in Neural Information Processing Systems},
  35:25792--25806, 2022.

\bibitem{ExploitingAndGalea2018}
Dieter Galea, Ivan Laponogov, and Kirill Veselkov.
\newblock Exploiting and assessing multi-source data for supervised biomedical
  named entity recognition.
\newblock {\em Bioinformatics}, 34:2474--2482, 7 2018.

\bibitem{RegelCorpusIGarda2022}
Samuele Garda, Freyda Lenihan-Geels, Sebastian Proft, Stefanie Hochmuth, Markus
  Sch\"{u}lke, Dominik Seelow, and Ulf Leser.
\newblock Regel corpus: identifying dna regulatory elements in the scientific
  literature.
\newblock {\em Database}, 2022, 6 2022.

\bibitem{garda2023belb}
Samuele Garda, Leon Weber-Genzel, Robert Martin, and Ulf Leser.
\newblock Belb: a biomedical entity linking benchmark.
\newblock {\em arXiv preprint arXiv:2308.11537}, 2023.

\bibitem{BelbABiomediGarda2023}
Samuele Garda, Leon Weber-Genzel, Robert Martin, and Ulf Leser.
\newblock Belb: a biomedical entity linking benchmark.
\newblock Aug 2023.

\bibitem{gerner2010linnaeus}
Martin Gerner, Goran Nenadic, and Casey~M Bergman.
\newblock Linnaeus: a species name identification system for biomedical
  literature.
\newblock {\em BMC bioinformatics}, 11(1):1--17, 2010.

\bibitem{giorgi2020towards}
John~M Giorgi and Gary~D Bader.
\newblock Towards reliable named entity recognition in the biomedical domain.
\newblock {\em Bioinformatics}, 36(1):280--286, 2020.

\bibitem{DBLP:journals/corr/abs-1811-04860}
Genevieve Gorrell, Xingyi Song, and Angus Roberts.
\newblock Bio-yodie: {A} named entity linking system for biomedical text.
\newblock {\em CoRR}, abs/1811.04860, 2018.

\bibitem{GuTCLULNGP22}
Yu~Gu, Robert Tinn, Hao Cheng, Michael Lucas, Naoto Usuyama, Xiaodong Liu,
  Tristan Naumann, Jianfeng Gao, and Hoifung Poon.
\newblock Domain-specific language model pretraining for biomedical natural
  language processing.
\newblock {\em {ACM} Trans. Comput. Heal.}, 3(1):2:1--2:23, 2022.

\bibitem{gurulingappa2010empirical}
Harsha Gurulingappa, Roman Klinger, Martin Hofmann-Apitius, and Juliane Fluck.
\newblock An empirical evaluation of resources for the identification of
  diseases and adverse effects in biomedical literature.
\newblock In {\em 2nd Workshop on Building and evaluating resources for
  biomedical text mining (7th edition of the Language Resources and Evaluation
  Conference)}, pages 15--22, 2010.

\bibitem{gyori_gilda_2022}
Benjamin~M Gyori, Charles~Tapley Hoyt, and Albert Steppi.
\newblock Gilda: biomedical entity text normalization with machine-learned
  disambiguation as a service.
\newblock {\em Bioinformatics Advances}, 2(1):vbac034, 2022.

\bibitem{HabibiWNWL17}
Maryam Habibi, Leon Weber, Mariana~L. Neves, David~Luis Wiegandt, and Ulf
  Leser.
\newblock Deep learning with word embeddings improves biomedical named entity
  recognition.
\newblock {\em Bioinform.}, 33(14):i37--i48, 2017.

\bibitem{hamosh2005online}
Ada Hamosh, Alan~F Scott, Joanna~S Amberger, Carol~A Bocchini, and Victor~A
  McKusick.
\newblock Online mendelian inheritance in man (omim), a knowledgebase of human
  genes and genetic disorders.
\newblock {\em Nucleic acids research},
  33(suppl\textbackslash{}\_1):D514--D517, 2005.

\bibitem{hochreiter1997long}
Sepp Hochreiter and J{\"u}rgen Schmidhuber.
\newblock Long short-term memory.
\newblock {\em Neural computation}, 9(8):1735--1780, 1997.

\bibitem{huang2020biomedical}
Ming-Siang Huang, Po-Ting Lai, Pei-Yen Lin, Yu-Ting You, Richard Tzong-Han
  Tsai, and Wen-Lian Hsu.
\newblock Biomedical named entity recognition and linking datasets: survey and
  our recent development.
\newblock {\em Briefings in Bioinformatics}, 21(6):2219--2238, 2020.

\bibitem{NlmChemANewIslama2021}
R~Islamaj, R~Leaman, S~Kim, D~Kwon, CH~Wei, DC~Comeau, Y~Peng, D~Cissel,
  C~Coss, C~Fisher, et~al.
\newblock Nlm-chem, a new resource for chemical entity recognition in pubmed
  full text literature.
\newblock {\em Scientific Data}, 8(1):91--91, 2021.

\bibitem{islamaj2021nlm}
Rezarta Islamaj, Robert Leaman, Sun Kim, Dongseop Kwon, Chih-Hsuan Wei,
  Donald~C Comeau, Yifan Peng, David Cissel, Cathleen Coss, Carol Fisher,
  et~al.
\newblock Nlm-chem, a new resource for chemical entity recognition in pubmed
  full text literature.
\newblock {\em Scientific data}, 8(1):91, 2021.

\bibitem{Islamaj2021}
Rezarta Islamaj, Chih-Hsuan Wei, David Cissel, Nicholas Miliaras, Olga
  Printseva, Oleg Rodionov, Keiko Sekiya, Janice Ward, and Zhiyong Lu.
\newblock Nlm-gene, a richly annotated gold standard dataset for gene entities
  that addresses ambiguity and multi-species gene recognition.
\newblock {\em Journal of biomedical informatics}, 118:103779, 2021.

\bibitem{JuMA18}
Meizhi Ju, Makoto Miwa, and Sophia Ananiadou.
\newblock A neural layered model for nested named entity recognition.
\newblock In Marilyn~A. Walker, Heng Ji, and Amanda Stent, editors, {\em
  Proceedings of the 2018 Conference of the North American Chapter of the
  Association for Computational Linguistics: Human Language Technologies,
  {NAACL-HLT} 2018, New Orleans, Louisiana, USA, June 1-6, 2018, Volume 1 (Long
  Papers)}, pages 1446--1459. Association for Computational Linguistics, 2018.

\bibitem{kaewphan2018wide}
Suwisa Kaewphan, Kai Hakala, Niko Miekka, Tapio Salakoski, and Filip Ginter.
\newblock Wide-scope biomedical named entity recognition and normalization with
  crfs, fuzzy matching and character level modeling.
\newblock {\em Database}, 2018:bay096, 2018.

\bibitem{AComprehensiveKartch2023}
David Kartchner, Jennifer Deng, Shubham Lohiya, Tejasri Kopparthi, Prasanth
  Bathala, Daniel Domingo-Fern{\'a}ndez, and Cassie Mitchell.
\newblock A comprehensive evaluation of biomedical entity linking models.
\newblock In Houda Bouamor, Juan Pino, and Kalika Bali, editors, {\em
  Proceedings of the 2023 Conference on Empirical Methods in Natural Language
  Processing}, pages 14462--14478, Singapore, December 2023. Association for
  Computational Linguistics.

\bibitem{KazamaMOT02}
Jun'ichi Kazama, Takaki Makino, Yoshihiro Ohta, and Jun'ichi Tsujii.
\newblock Tuning support vector machines for biomedical named entity
  recognition.
\newblock In Stephen Johnson, editor, {\em Proceedings of the {ACL} 2002
  Workshop on Natural Language Processing in the Biomedical Domain, July 11,
  2002, University of Pennsylvania, Philadelphia, PA, {USA}}, pages 1--8.
  {ACL}, 2002.

\bibitem{kim2019corpus}
Baeksoo Kim, Wonjun Choi, and Hyunju Lee.
\newblock A corpus of plant--disease relations in the biomedical domain.
\newblock {\em PLoS One}, 14(8):e0221582, 2019.

\bibitem{kim2019neural}
Donghyeon Kim, Jinhyuk Lee, Chan~Ho So, Hwisang Jeon, Minbyul Jeong, Yonghwa
  Choi, Wonjin Yoon, Mujeen Sung, and Jaewoo Kang.
\newblock A neural named entity recognition and multi-type normalization tool
  for biomedical text mining.
\newblock {\em IEEE Access}, 7:73729--73740, 2019.

\bibitem{kocaman2021biomedical}
Veysel Kocaman and David Talby.
\newblock Biomedical named entity recognition at scale.
\newblock In {\em International Conference on Pattern Recognition}, pages
  635--646, 2021.

\bibitem{kolarik2008chemical}
Corinna Kol{\'a}rik, Roman Klinger, Christoph~M Friedrich, Martin
  Hofmann-Apitius, and Juliane Fluck.
\newblock Chemical names: terminological resources and corpora annotation.
\newblock In {\em Workshop on Building and evaluating resources for biomedical
  text mining (6th edition of the Language Resources and Evaluation
  Conference)}, volume~36, 2008.

\bibitem{kormilitzin2021med7}
Andrey Kormilitzin, Nemanja Vaci, Qiang Liu, and Alejo Nevado-Holgado.
\newblock Med7: A transferable clinical natural language processing model for
  electronic health records.
\newblock {\em Artificial Intelligence in Medicine}, 118:102086, 2021.

\bibitem{KorvigoHZS18}
Ilia Korvigo, Maxim Holmatov, Anatolii Zaikovskii, and Mikhail Skoblov.
\newblock Putting hands to rest: efficient deep {CNN-RNN} architecture for
  chemical named entity recognition with no hand-crafted rules.
\newblock {\em J. Cheminformatics}, 10(1):28, 2018.

\bibitem{EvaluationMeasKosmop2015}
Aris Kosmopoulos, Ioannis Partalas, Eric Gaussier, Georgios Paliouras, and Ion
  Androutsopoulos.
\newblock Evaluation measures for hierarchical classification: a unified view
  and novel approaches.
\newblock {\em Data Mining and Knowledge Discovery}, 29:820--865, 5 2015.

\bibitem{MultiDomainClKralje2021}
Zeljko Kraljevic, Thomas Searle, Anthony Shek, Lukasz Roguski, Kawsar Noor,
  Daniel Bean, Aurelie Mascio, Leilei Zhu, Amos~A. Folarin, Angus Roberts,
  Rebecca Bendayan, Mark~P. Richardson, Robert Stewart, Anoop~D. Shah,
  Wai~Keong Wong, Zina Ibrahim, James~T. Teo, and Richard~J.B. Dobson.
\newblock Multi-domain clinical natural language processing with medcat: The
  medical concept annotation toolkit.
\newblock {\em Artificial Intelligence in Medicine}, 117:102083, 7 2021.

\bibitem{krallinger2015chemdner}
Martin Krallinger, Obdulia Rabal, Florian Leitner, Miguel Vazquez, David
  Salgado, Zhiyong Lu, Robert Leaman, Yanan Lu, Donghong Ji, Daniel~M Lowe,
  et~al.
\newblock The chemdner corpus of chemicals and drugs and its annotation
  principles.
\newblock {\em Journal of cheminformatics}, 7(1):1--17, 2015.

\bibitem{lample2016neural}
Guillaume Lample, Miguel Ballesteros, Sandeep Subramanian, Kazuya Kawakami, and
  Chris Dyer.
\newblock Neural architectures for named entity recognition.
\newblock In {\em Proceedings of the 2016 Conference of the North American
  Chapter of the Association for Computational Linguistics: Human Language
  Technologies}, pages 260--270, 2016.

\bibitem{leaman_banner_2007}
Robert Leaman and Graciela Gonzalez.
\newblock {BANNER}: {AN} {EXECUTABLE} {SURVEY} {OF} {ADVANCES} {IN}
  {BIOMEDICAL} {NAMED} {ENTITY} {RECOGNITION}.
\newblock In {\em Biocomputing 2008}, pages 652--663. {WORLD} {SCIENTIFIC},
  2007.

\bibitem{ChemicalIdentiLeaman2023}
Robert Leaman, Rezarta Islamaj, Virginia Adams, Mohammed~A Alliheedi,
  Jo\~{a}o~Rafael Almeida, Rui Antunes, Robert Bevan, Yung-Chun Chang, Arslan
  Erdengasileng, Matthew Hodgskiss, Ryuki Ida, Hyunjae Kim, Keqiao Li, Robert~E
  Mercer, Lukr\'{e}cia Mertov\'{a}, Ghadeer Mobasher, Hoo-Chang Shin, Mujeen
  Sung, Tomoki Tsujimura, Wen-Chao Yeh, and Zhiyong Lu.
\newblock Chemical identification and indexing in full-text articles: an
  overview of the nlm-chem track at biocreative vii.
\newblock {\em Database}, 2023, 3 2023.

\bibitem{leaman2013dnorm}
Robert Leaman, Rezarta Islamaj~Do{\u{g}}an, and Zhiyong Lu.
\newblock Dnorm: disease name normalization with pairwise learning to rank.
\newblock {\em Bioinformatics}, 29(22):2909--2917, 2013.

\bibitem{Leaman2016}
Robert Leaman and Zhiyong Lu.
\newblock Taggerone: joint named entity recognition and normalization with
  semi-markov models.
\newblock {\em Bioinformatics}, 32(18):2839\textendash{}2846, 2016.

\bibitem{leaman2015tmchem}
Robert Leaman, Chih-Hsuan Wei, and Zhiyong Lu.
\newblock tmchem: a high performance approach for chemical named entity
  recognition and normalization.
\newblock {\em Journal of Cheminformatics}, 7(S1):S3, 2015.

\bibitem{LeeYKKKSK20}
Jinhyuk Lee, Wonjin Yoon, Sungdong Kim, Donghyeon Kim, Sunkyu Kim, Chan~Ho So,
  and Jaewoo Kang.
\newblock Biobert: a pre-trained biomedical language representation model for
  biomedical text mining.
\newblock {\em Bioinform.}, 36(4):1234--1240, 2020.

\bibitem{leser2005makes}
Ulf Leser and J{\"o}rg Hakenberg.
\newblock What makes a gene name? named entity recognition in the biomedical
  literature.
\newblock {\em Briefings in bioinformatics}, 6(4):357--369, 2005.

\bibitem{Li2016a}
Jiao Li, Yueping Sun, Robin~J. Johnson, Daniela Sciaky, Chih-Hsuan Wei, Robert
  Leaman, Allan~Peter Davis, Carolyn~J. Mattingly, Thomas~C. Wiegers, and
  Zhiyong Lu.
\newblock Biocreative v cdr task corpus: a resource for chemical disease
  relation extraction.
\newblock {\em Database}, 2016(baw068), 2016.

\bibitem{li2023rethinking}
Jing Li, Yequan Wang, Shuai Zhang, and Min Zhang.
\newblock Rethinking document-level relation extraction: A reality check.
\newblock {\em arXiv preprint arXiv:2306.08953}, 2023.

\bibitem{lipscomb2000medical}
Carolyn~E Lipscomb.
\newblock Medical subject headings (mesh).
\newblock {\em Bulletin of the Medical Library Association}, 88(3):265, 2000.

\bibitem{Liu2021a}
Fangyu Liu, Ehsan Shareghi, Zaiqiao Meng, Marco Basaldella, and Nigel Collier.
\newblock Self-alignment pretraining for biomedical entity representations.
\newblock In {\em Proceedings of the 2021 Conference of the North American
  Chapter of the Association for Computational Linguistics: Human Language
  Technologies}, page 4228\textendash{}4238. Association for Computational
  Linguistics, 2021.

\bibitem{liu2019roberta}
Yinhan Liu, Myle Ott, Naman Goyal, Jingfei Du, Mandar Joshi, Danqi Chen, Omer
  Levy, Mike Lewis, Luke Zettlemoyer, and Veselin Stoyanov.
\newblock Roberta: A robustly optimized bert pretraining approach.
\newblock {\em arXiv preprint arXiv:1907.11692}, 2019.

\bibitem{luo2022biored}
Ling Luo, Po-Ting Lai, Chih-Hsuan Wei, Cecilia~N Arighi, and Zhiyong Lu.
\newblock Biored: a rich biomedical relation extraction dataset.
\newblock {\em Briefings in Bioinformatics}, 23(5):bbac282, 2022.

\bibitem{luo2023aioner}
Ling Luo, Chih-Hsuan Wei, Po-Ting Lai, Robert Leaman, Qingyu Chen, and Zhiyong
  Lu.
\newblock Aioner: all-in-one scheme-based biomedical named entity recognition
  using deep learning.
\newblock {\em Bioinformatics}, 39(5):btad310, 2023.

\bibitem{LyuCRJ17}
Chen Lyu, Bo~Chen, Yafeng Ren, and Donghong Ji.
\newblock Long short-term memory {RNN} for biomedical named entity recognition.
\newblock {\em {BMC} Bioinform.}, 18(1):462, 2017.

\bibitem{mohanmedmentions}
Sunil Mohan and Donghui Li.
\newblock Medmentions: A large biomedical corpus annotated with umls concepts.
\newblock In {\em In Proceedings of the 2019 Conference on Automated Knowledge
  Base Construction (AKBC 2019)}, 2019.

\bibitem{OverviewOfBioMorgan2008}
Alexander~A Morgan, Zhiyong Lu, Xinglong Wang, Aaron~M Cohen, Juliane Fluck,
  Patrick Ruch, Anna Divoli, Katrin Fundel, Robert Leaman, J\"{o}rg Hakenberg,
  Chengjie Sun, Heng-hui Liu, Rafael Torres, Michael Krauthammer, William~W
  Lau, Hongfang Liu, Chun-Nan Hsu, Martijn Schuemie, K~Bretonnel Cohen, and
  Lynette Hirschman.
\newblock Overview of biocreative ii gene normalization.
\newblock {\em Genome Biology}, 9:S3, 2008.

\bibitem{Bern2AnAdvanMujeen}
Sung Mujeen, Jeong Minbyul, Choi Yonghwa, Kim Donghyeon, Lee Jinhyuk, and Kang
  Jaewoo.
\newblock Bern2: an advanced neural biomedical named entity recognition and
  normalization tool.
\newblock {\em Bioinformatics}, 38, 2022.

\bibitem{neumann2019scispacy}
Mark Neumann, Daniel King, Iz~Beltagy, and Waleed Ammar.
\newblock Scispacy: fast and robust models for biomedical natural language
  processing.
\newblock {\em arXiv preprint arXiv:1902.07669}, 2019.

\bibitem{ouyangTrainingLanguageModels2022}
Long Ouyang, Jeffrey Wu, Xu~Jiang, Diogo Almeida, Carroll~L. Wainwright, Pamela
  Mishkin, Chong Zhang, Sandhini Agarwal, Katarina Slama, Alex Ray, John
  Schulman, Jacob Hilton, Fraser Kelton, Luke Miller, Maddie Simens, Amanda
  Askell, Peter Welinder, Paul~F. Christiano, Jan Leike, and Ryan Lowe.
\newblock Training language models to follow instructions with human feedback.
\newblock In {\em NeurIPS}, 2022.

\bibitem{pafilis2013species}
Evangelos Pafilis, Sune~P Frankild, Lucia Fanini, Sarah Faulwetter, Christina
  Pavloudi, Aikaterini Vasileiadou, Christos Arvanitidis, and Lars~Juhl Jensen.
\newblock The species and organisms resources for fast and accurate
  identification of taxonomic names in text.
\newblock {\em PloS one}, 8(6):e65390, 2013.

\bibitem{Page1999ThePC}
Lawrence Page, Sergey Brin, Rajeev Motwani, and Terry Winograd.
\newblock The pagerank citation ranking : Bringing order to the web.
\newblock In {\em The Web Conference}, 1999.

\bibitem{PengYL19}
Yifan Peng, Shankai Yan, and Zhiyong Lu.
\newblock Transfer learning in biomedical natural language processing: An
  evaluation of {BERT} and elmo on ten benchmarking datasets.
\newblock In Dina Demner{-}Fushman, Kevin~Bretonnel Cohen, Sophia Ananiadou,
  and Junichi Tsujii, editors, {\em Proceedings of the 18th BioNLP Workshop and
  Shared Task, BioNLP@ACL 2019, Florence, Italy, August 1, 2019}, pages 58--65.
  Association for Computational Linguistics, 2019.

\bibitem{TransferLearniPeng2019}
Yifan Peng, Shankai Yan, and Zhiyong Lu.
\newblock Transfer learning in biomedical natural language processing: An
  evaluation of bert and elmo on ten benchmarking datasets.
\newblock In {\em Proceedings of the 18th BioNLP Workshop and Shared Task},
  pages 58--65, Florence, Italy, August 2019. Association for Computational
  Linguistics.

\bibitem{perera2020named}
Nadeesha Perera, Matthias Dehmer, and Frank Emmert-Streib.
\newblock Named entity recognition and relation detection for biomedical
  information extraction.
\newblock {\em Frontiers in cell and developmental biology}, page 673, 2020.

\bibitem{Phan2019}
Minh~C. Phan, Aixin Sun, and Yi~Tay.
\newblock Robust representation learning of biomedical names.
\newblock In {\em Proceedings of the 57th Annual Meeting of the Association for
  Computational Linguistics}, page 3275\textendash{}3285. Association for
  Computational Linguistics, 2019.

\bibitem{pyysalo2013overview}
Sampo Pyysalo, Tomoko Ohta, and Sophia Ananiadou.
\newblock Overview of the cancer genetics (cg) task of bionlp shared task 2013.
\newblock In {\em Proceedings of the BioNLP Shared Task 2013 Workshop}, pages
  58--66, 2013.

\bibitem{LinkingChemicaRuas2020}
Pedro Ruas, Andre Lamurias, and Francisco~M. Couto.
\newblock Linking chemical and disease entities to ontologies by integrating
  pagerank with extracted relations from literature.
\newblock {\em Journal of Cheminformatics}, 12, 12 2020.

\bibitem{ruas2023lasige}
Pedro Ruas, Diana~F Sousa, Andr{\'e} Neves, Carlos Cruz, and Francisco~M Couto.
\newblock Lasige and unicage solution to the nasa litcoin nlp competition.
\newblock {\em arXiv preprint arXiv:2308.05609}, 2023.

\bibitem{sanger2021large}
Mario S{\"a}nger and Ulf Leser.
\newblock Large-scale entity representation learning for biomedical
  relationship extraction.
\newblock {\em Bioinformatics}, 37(2):236--242, 2021.

\bibitem{MayoClinicalTSavova2010}
Guergana~K Savova, James~J Masanz, Philip~V Ogren, Jiaping Zheng, Sunghwan
  Sohn, Karin~C Kipper-Schuler, and Christopher~G Chute.
\newblock Mayo clinical text analysis and knowledge extraction system (ctakes):
  architecture, component evaluation and applications.
\newblock {\em Journal of the American Medical Informatics Association},
  17:507--513, 9 2010.

\bibitem{TheNcbiTaxonoScott2012}
Federhen Scott.
\newblock The ncbi taxonomy database.
\newblock {\em Nucleic Acids Research}, 40:D136--D143, 1 2012.

\bibitem{Settles04}
Burr Settles.
\newblock Biomedical named entity recognition using conditional random fields
  and rich feature sets.
\newblock In Nigel Collier, Patrick Ruch, and Adeline Nazarenko, editors, {\em
  Proceedings of the International Joint Workshop on Natural Language
  Processing in Biomedicine and its Applications, NLPBA/BioNLP 2004, Geneva,
  Switzerland, August 28-29, 2004}, 2004.

\bibitem{ShenZZST03}
Dan Shen, Jie Zhang, Guodong Zhou, Jian Su, and Chew~Lim Tan.
\newblock Effective adaptation of hidden markov model-based named entity
  recognizer for biomedical domain.
\newblock In {\em Proceedings of the Workshop on Natural Language Processing in
  Biomedicine, BioNLP@ACL 2003, Sapporo, Japan, July 2003}, pages 49--56, 2003.

\bibitem{smith2008overview}
Larry Smith, Lorraine~K Tanabe, Cheng-Ju Kuo, I~Chung, Chun-Nan Hsu, Yu-Shi
  Lin, Roman Klinger, Christoph~M Friedrich, Kuzman Ganchev, Manabu Torii,
  et~al.
\newblock Overview of biocreative ii gene mention recognition.
\newblock {\em Genome biology}, 9(2):1--19, 2008.

\bibitem{AbbreviationDeSohn2008}
Sunghwan Sohn, Donald~C Comeau, Won Kim, and W~John Wilbur.
\newblock Abbreviation definition identification based on automatic precision
  estimates.
\newblock {\em BMC Bioinformatics}, 9, 12 2008.

\bibitem{soldaini2016quickumls}
Luca Soldaini and Nazli Goharian.
\newblock Quickumls: a fast, unsupervised approach for medical concept
  extraction.
\newblock In {\em MedIR workshop, sigir}, pages 1--4, 2016.

\bibitem{SongLLZ21}
Bosheng Song, Fen Li, Yuansheng Liu, and Xiangxiang Zeng.
\newblock Deep learning methods for biomedical named entity recognition: a
  survey and qualitative comparison.
\newblock {\em Briefings Bioinform.}, 22(6), 2021.

\bibitem{song2021deep}
Bosheng Song, Fen Li, Yuansheng Liu, and Xiangxiang Zeng.
\newblock Deep learning methods for biomedical named entity recognition: a
  survey and qualitative comparison.
\newblock {\em Briefings in Bioinformatics}, 22(6):bbab282, 2021.

\bibitem{soysal_clamp_2018}
Ergin Soysal, Jingqi Wang, Min Jiang, Yonghui Wu, Serguei Pakhomov, Hongfang
  Liu, and Hua Xu.
\newblock {CLAMP} – a toolkit for efficiently building customized clinical
  natural language processing pipelines.
\newblock {\em Journal of the American Medical Informatics Association},
  25(3):331--336, 2018.

\bibitem{su2022deep}
Yansen Su, Minglu Wang, Pengpeng Wang, Chunhou Zheng, Yuansheng Liu, and
  Xiangxiang Zeng.
\newblock Deep learning joint models for extracting entities and relations in
  biomedical: a survey and comparison.
\newblock {\em Briefings in Bioinformatics}, 23(6):bbac342, 2022.

\bibitem{Sung2020}
Mujeen Sung, Hwisang Jeon, Jinhyuk Lee, and Jaewoo Kang.
\newblock Biomedical entity representations with synonym marginalization.
\newblock In {\em Proceedings of the 58th Annual Meeting of the Association for
  Computational Linguistics}, page 3641\textendash{}3650. Association for
  Computational Linguistics, 2020.

\bibitem{tikk2010comprehensive}
Domonkos Tikk, Philippe Thomas, Peter Palaga, J{\"o}rg Hakenberg, and Ulf
  Leser.
\newblock A comprehensive benchmark of kernel methods to extract
  protein--protein interactions from literature.
\newblock {\em PLoS computational biology}, 6(7):e1000837, 2010.

\bibitem{trieuDeepEventMineEndtoendNeural2020}
Hai-Long Trieu, Thy~Thy Tran, Khoa N.~A. Duong, Anh Nguyen, Makoto Miwa, and
  Sophia Ananiadou.
\newblock {{DeepEventMine}}: End-to-end neural nested event extraction from
  biomedical texts.
\newblock {\em Bioinformatics (Oxford, England)}, 36(19):4910--4917, December
  2020.

\bibitem{EndToEndBiomUjiie2021}
Shogo Ujiie, Hayate Iso, Shuntaro Yada, Shoko Wakamiya, and Eiji Aramaki.
\newblock End-to-end biomedical entity linking with span-based dictionary
  matching.
\newblock In {\em Proceedings of the 20th Workshop on Biomedical Language
  Processing}, pages 162--167, Online, June 2021. Association for Computational
  Linguistics.

\bibitem{ushio_t-ner_2021}
Asahi Ushio and Jose Camacho-Collados.
\newblock T-{NER}: An all-round python library for transformer-based named
  entity recognition.
\newblock In {\em Proceedings of the 16th Conference of the European Chapter of
  the Association for Computational Linguistics: System Demonstrations}, pages
  53--62, 2021.

\bibitem{CrossDomainDaVarma2021}
Maya Varma, Laurel Orr, Sen Wu, Megan Leszczynski, Xiao Ling, and Christopher
  R\'{e}.
\newblock Cross-domain data integration for named entity disambiguation in
  biomedical text.
\newblock {\em Findings of the Association for Computational Linguistics: EMNLP
  2021}, 11 2021.

\bibitem{Vasilevsky2020}
N~Vasilevsky, S~Essaid, N~Matentzoglu, N~Harris, M~Haendel, P~Robinson, and
  C~Mungall.
\newblock Mondo disease ontology: harmonizing disease concepts across the
  world.
\newblock In {\em CEUR Workshop Proceedings}, volume 2807, page
  CEUR\textendash{}WS, 2020.

\bibitem{wang2023pre}
Benyou Wang, Qianqian Xie, Jiahuan Pei, Zhihong Chen, Prayag Tiwari, Zhao Li,
  and Jie Fu.
\newblock Pre-trained language models in biomedical domain: A systematic
  survey.
\newblock {\em ACM Computing Surveys}, 56(3):1--52, 2023.

\bibitem{wang-etal-2020-biomedical}
Xing~David Wang, Leon Weber, and Ulf Leser.
\newblock Biomedical event extraction as multi-turn question answering.
\newblock In {\em Proceedings of the 11th International Workshop on Health Text
  Mining and Information Analysis}, pages 88--96, Online, November 2020.
  Association for Computational Linguistics.

\bibitem{wang2018comparative}
Xu~Wang, Chen Yang, and Renchu Guan.
\newblock A comparative study for biomedical named entity recognition.
\newblock {\em International Journal of Machine Learning and Cybernetics},
  9:373--382, 2018.

\bibitem{weber2020huner}
Leon Weber, Jannes M{\"u}nchmeyer, Tim Rockt{\"a}schel, Maryam Habibi, and Ulf
  Leser.
\newblock Huner: improving biomedical ner with pretraining.
\newblock {\em Bioinformatics}, 36(1):295--302, 2020.

\bibitem{weber2022chemical}
Leon Weber, Mario S{\"a}nger, Samuele Garda, Fabio Barth, Christoph Alt, and
  Ulf Leser.
\newblock Chemical--protein relation extraction with ensembles of carefully
  tuned pretrained language models.
\newblock {\em Database}, 2022:baac098, 2022.

\bibitem{weber2021hunflair}
Leon Weber, Mario S{\"a}nger, Jannes M{\"u}nchmeyer, Maryam Habibi, Ulf Leser,
  and Alan Akbik.
\newblock Hunflair: an easy-to-use tool for state-of-the-art biomedical named
  entity recognition.
\newblock {\em Bioinformatics}, 37(17):2792--2794, 2021.

\bibitem{weber2020pedl}
Leon Weber, Kirsten Thobe, Oscar~Arturo Migueles~Lozano, Jana Wolf, and Ulf
  Leser.
\newblock Pedl: extracting protein--protein associations using deep language
  models and distant supervision.
\newblock {\em Bioinformatics}, 36(Supplement\_1):i490--i498, 2020.

\bibitem{weiPubTatorCentralAutomated2019}
Chih-Hsuan Wei, Alexis Allot, Robert Leaman, and Zhiyong Lu.
\newblock {{PubTator}} central: Automated concept annotation for biomedical
  full text articles.
\newblock {\em Nucleic Acids Research}, 47(W1):W587--W593, July 2019.

\bibitem{Wei2022}
Chih-Hsuan Wei, Alexis Allot, Kevin Riehle, Aleksandar Milosavljevic, and
  Zhiyong Lu.
\newblock tmvar 3.0: an improved variant concept recognition and normalization
  tool.
\newblock {\em Bioinformatics}, 38(18):4449\textendash{}4451, 2022.

\bibitem{CrossSpeciesGWeiC2011}
Chih-Hsuan Wei and Hung-Yu Kao.
\newblock Cross-species gene normalization by species inference.
\newblock {\em BMC Bioinformatics}, 12, 12 2011.

\bibitem{Sr4gnASpecieWeiC2012}
Chih-Hsuan Wei, Hung-Yu Kao, and Zhiyong Lu.
\newblock Sr4gn: A species recognition software tool for gene normalization.
\newblock {\em PLOS ONE}, 7:e38460, 6 2012.

\bibitem{Wei2015}
Chih-Hsuan Wei, Hung-Yu Kao, and Zhiyong Lu.
\newblock Gnormplus: An integrative approach for tagging genes, gene families,
  and protein domains.
\newblock {\em BioMed Research International}, 2015:e918710, 2015.

\bibitem{wei2018tmvar}
Chih-Hsuan Wei, Lon Phan, Juliana Feltz, Rama Maiti, Tim Hefferon, and Zhiyong
  Lu.
\newblock tmvar 2.0: integrating genomic variant information from literature
  with dbsnp and clinvar for precision medicine.
\newblock {\em Bioinformatics}, 34(1):80--87, 2018.

\bibitem{wen2019desiderata}
Andrew Wen, Sunyang Fu, Sungrim Moon, Mohamed El~Wazir, Andrew Rosenbaum,
  Vinod~C Kaggal, Sijia Liu, Sunghwan Sohn, Hongfang Liu, and Jungwei Fan.
\newblock Desiderata for delivering nlp to accelerate healthcare ai advancement
  and a mayo clinic nlp-as-a-service implementation.
\newblock {\em NPJ digital medicine}, 2(1):130, 2019.

\bibitem{BiomedicalConcYanC2021}
Cheng Yan, Yuanzhe Zhang, Kang Liu, Jun Zhao, Yafei Shi, and Shengping Liu.
\newblock Biomedical concept normalization by leveraging hypernyms.
\newblock In {\em Proceedings of the 2021 Conference on Empirical Methods in
  Natural Language Processing}, pages 3512--3517, Online and Punta Cana,
  Dominican Republic, November 2021. Association for Computational Linguistics.

\bibitem{YasunagaLL22}
Michihiro Yasunaga, Jure Leskovec, and Percy Liang.
\newblock Linkbert: Pretraining language models with document links.
\newblock In Smaranda Muresan, Preslav Nakov, and Aline Villavicencio, editors,
  {\em Proceedings of the 60th Annual Meeting of the Association for
  Computational Linguistics (Volume 1: Long Papers), {ACL} 2022, Dublin,
  Ireland, May 22-27, 2022}, pages 8003--8016. Association for Computational
  Linguistics, 2022.

\bibitem{GenerativeBiomYuan2022}
Hongyi Yuan, Zheng Yuan, and Sheng Yu.
\newblock Generative biomedical entity linking via knowledge base-guided
  pre-training and synonyms-aware fine-tuning.
\newblock In {\em Proceedings of the 2022 Conference of the North American
  Chapter of the Association for Computational Linguistics: Human Language
  Technologies}, pages 4038--4048, Seattle, United States, July 2022.
  Association for Computational Linguistics.

\bibitem{zeng2006extracting}
Qing~T Zeng, Sergey Goryachev, Scott Weiss, Margarita Sordo, Shawn~N Murphy,
  and Ross Lazarus.
\newblock Extracting principal diagnosis, co-morbidity and smoking status for
  asthma research: evaluation of a natural language processing system.
\newblock {\em BMC medical informatics and decision making}, 6(1):1--9, 2006.

\bibitem{zhang2022knowledge}
Sheng Zhang, Hao Cheng, Shikhar Vashishth, Cliff Wong, Jinfeng Xiao, Xiaodong
  Liu, Tristan Naumann, Jianfeng Gao, and Hoifung Poon.
\newblock Knowledge-rich self-supervision for biomedical entity linking.
\newblock In {\em Findings of the Association for Computational Linguistics:
  EMNLP 2022}, pages 868--880, 2022.

\bibitem{zhang_biomedical_2021}
Yuhao Zhang, Yuhui Zhang, Peng Qi, Christopher~D Manning, and Curtis~P
  Langlotz.
\newblock Biomedical and clinical english model packages for the stanza python
  {NLP} library.
\newblock {\em Journal of the American Medical Informatics Association},
  28(9):1892--1899, 2021.

\end{thebibliography}


%

\clearpage

\begin{appendices}
	\section{Overview over all surveyed tools}\label{app:tools}

\afterpage{%
	\begin{landscape}
		\begin{table}[thbp]
			\centering
			\caption{List of entity extraction tools identified by our literature review.
				For the non-selected tools we report the unfullfilled criteria (see \ref{sec:bio_ee_tools}) disqualifying from our evaluation.
				Abbreviations used in the table are:
				\enquote{RB} for rule-based systems,
				\enquote{ML} denote all standard machine learning approaches (e.g. Support Vector Machines) distinguished from neural-network based (\enquote{N}) ones.
				Entity types we consider are: genes (Ge), species (Sp), disease (Di), chemical (Ch), cell line (Cl), variant (Va).
				We also include a miscellaneous category (Misc.) if the tools supports additional (clinical) entity types.
				\cmark ~denotes support for both NER and NEN,
				(\cmark) ~denotes support for either NER or NEN only.
                Last update refers to the date when the corresponding GitHub repository was changed last.
				Citation count refers to the number of Google scholar citations as of the 10/01/2024.
				$\dagger$ Requires UMLS, which has its own license
				$\ddag$ Commercial}
			\begin{adjustbox}{width=1.35\textwidth}
				\begin{tabular}{lllccccccllllllll}
					\toprule
					\textbf{Tools}                  & \textbf{Ref.}                                                   &\textbf{API}                 & \textbf{Ge}       & \textbf{Sp}       & \textbf{Di}       & \textbf{Ch}       & \textbf{Cl}      & \textbf{Va}     & \textbf{Misc.}  & \textbf{NER}         & \textbf{NEN}     & \textbf{Pub. year} & \textbf{Last update}                               & \textbf{Citations} & \textbf{License}                                                & \textbf{Exclusion}   \\                         
					\midrule                                                                                                                                                                                                                                                                                                                                                          
					\textbf{Non-selected}  &                                                        &                     &          &          &          &          &          &          &          &             &         &        &                       &           &                                                        & \textcolor{blue}{C}\ref{crit:entext}, \textcolor{blue}{C}\ref{crit:entity}                                          \\
					\quad BANNER           & \cite{leaman_banner_2007}                              & Java                & (\cmark) &          & (\cmark) &          &          &          &          & ML          & -       &  2007  & -                     & 635       & CPL                                                    & \textcolor{blue}{C}\ref{crit:entext}                                                                                \\
					\quad EasyNER          & \cite{ahmed2023easyner}                                & Python              & (\cmark) & (\cmark) & (\cmark) & (\cmark) & (\cmark) &          &          & N           & -       &  2023  & 12/2023               & 1         & Apache-2.0 license                                     & \textcolor{blue}{C}\ref{crit:entext}, \textcolor{blue}{C}\ref{crit:ml_based}, \textcolor{blue}{C}\ref{crit:entity}  \\
					\quad Gimli            & \cite{campos_gimli_2013}                               & Java                & (\cmark) &          &          &          & (\cmark) &          &          & RB + ML     & -       &  2013  & -                     & 136       & -                                                      & \textcolor{blue}{C}\ref{crit:entext}, \textcolor{blue}{C}\ref{crit:entity}                                          \\
					\quad Med7             & \cite{kormilitzin2021med7}                             & Python              &          &          &          &  		  &          &          & (\cmark) & N + ML      & -       &  2021  & 12/2021               & 86        & Apache 2.0                                             & \textcolor{blue}{C}\ref{crit:entext}, \textcolor{blue}{C}\ref{crit:ml_based}, \textcolor{blue}{C}\ref{crit:entity}  \\ 
					\quad MedTagger        & \cite{wen2019desiderata}                               & Java                &          &          &  		   &          &          &          & (\cmark) & RB          & -       &  2019  & 10/2023               & 77        & Apache-2.0 license                                     & \textcolor{blue}{C}\ref{crit:entext}, \textcolor{blue}{C}\ref{crit:entity}                                          \\
					\quad NeuroNER         & \cite{dernoncourt2017neuroner}                         & Python              &          &          &          &          &          &          & (\cmark) & N           & -       &  2017  & 10/2019               & 241       & MIT license                                            & \textcolor{blue}{C}\ref{crit:entext}                                                                                \\
					\quad SparkNLP         & \cite{kocaman2021biomedical}                           & Python,R,Scala,Java & (\cmark) & (\cmark) & (\cmark) & (\cmark) & (\cmark) &          & (\cmark) & N           & -       &  2021  & 01/2024               & 54        & Apache 2.0 $\ddag$                                     & \textcolor{blue}{C}\ref{crit:entext}                                                                                \\
					\quad Stanza           & \cite{zhang_biomedical_2021}                           & Python              & (\cmark) & (\cmark) & (\cmark) & (\cmark) & (\cmark) &          & (\cmark) & N           & -       &  2021  & 12/2023               & 118       & Apache 2.0                                             & \textcolor{blue}{C}\ref{crit:entext}, \textcolor{blue}{C}\ref{crit:entity}                                          \\
					\quad TNER             & \cite{ushio_t-ner_2021}                                & Python              & (\cmark) &          &          &          & (\cmark) &          &          & N           & -       &  2021  & 05/2023               & 51        & MIT license                                            & \textcolor{blue}{C}\ref{crit:entext}, \textcolor{blue}{C}\ref{crit:entity}                                          \\
					\quad ChatGPT          & \cite{ouyangTrainingLanguageModels2022}                & Web/REST            &          &          &          &          &          &          &          & N           & -       &  2022  & -                     & 3952      & ToS                                                    & \textcolor{blue}{C}\ref{crit:entext}, \textcolor{blue}{C}\ref{crit:entity}                                          \\
					\quad Gilda            & \cite{gyori_gilda_2022}                                & Python/REST         & (\cmark) &          &          &  		  &          &          & (\cmark) & -           & RB + ML &  2022  & 12/2023               & 8         & BSD2                                                   & \textcolor{blue}{C}\ref{crit:ml_based}                                                                              \\
					\quad Bio-YODIE        & \cite{DBLP:journals/corr/abs-1811-04860}               & Java                &          & \cmark   & \cmark   & \cmark   &          &          &          & RB + N      & RB + N  &  2018  & 10/2019               & 22        & GNU Affero General Public Licence                      & \textcolor{blue}{C}\ref{crit:ml_based}                                                                              \\
					\quad CLAMP            & \cite{soysal_clamp_2018}                               & Java                & \cmark   & \cmark   & \cmark   & \cmark   & \cmark   & \cmark   & \cmark   & ML + RB     & RB      &  2018  & - 					& 358       & Free for research use                                  & \textcolor{blue}{C}\ref{crit:ml_based}, \textcolor{blue}{C}\ref{crit:entity}, \textcolor{blue}{C}\ref{crit:license}                                        \\
					\quad cTAKES           & \cite{MayoClinicalTSavova2010}                         & Java                & \cmark   & \cmark   & \cmark   &          & \cmark   & \cmark   & \cmark   & RB          & RB      &  2010  & -                     & 2148      & Apache License V2.0 $\dagger$                          & \textcolor{blue}{C}\ref{crit:ml_based}, \textcolor{blue}{C}\ref{crit:entity}, \textcolor{blue}{C}\ref{crit:license}                                        \\
					\quad HITEx            & \cite{zeng2006extracting}                              & Java                &          &          &          &          &          &          & \cmark   & ML + RB     & RB      &  2006  & -                     & 433       & Open source, i2b2 $\dagger$ 							 & \textcolor{blue}{C}\ref{crit:ml_based}                                                                              \\
					\quad MedCat           & \cite{MultiDomainClKralje2021}                         & Python              &  		 & \cmark   & \cmark   & \cmark   &          &          & \cmark   & RB          & RB + N  &  2021  & 12/2023               & 107       & Elastic License 2.0                                    & \textcolor{blue}{C}\ref{crit:ml_based}                                                                              \\
					\quad MedSpaCy         & \cite{eyre2021launching}                               & Python              &          &          &          &  		  &          &          & \cmark   & N + ML + RB & RB      &  2021  & 01/2024               & 51        & MIT license                                            & \textcolor{blue}{C}\ref{crit:ml_based}, \textcolor{blue}{C}\ref{crit:entity}                                        \\
					\quad MetaMap Lite     & \cite{demner-fushman_metamap_2017,aronson2010overview} & Java/Web+REST       & \cmark   & \cmark   & \cmark   & \cmark   & \cmark   & \cmark   & \cmark   & RB          & RB      &  2015  & 07/2022               & 172       & BSD license $\dagger$                                  & \textcolor{blue}{C}\ref{crit:ml_based}, , \textcolor{blue}{C}\ref{crit:license}                                                                              \\
					\quad Neji             & \cite{AModularFrameCampos2013}                         & Java                & \cmark   & \cmark   & \cmark   & \cmark   & \cmark   &          &          & RB + ML     & RB      &  2013  & 05/2017               & 93        & CC BY-NC-SA 3.0                                        & \textcolor{blue}{C}\ref{crit:ml_based}                                                                              \\
					\quad OnTheFly2.0      & \cite{Onthefly20ABaltou2021}                           & Web                 & \cmark   & \cmark   & \cmark   &          &          &          & \cmark   & RB          & RB      &  2021  & 07/2022               & 14        & GPLv3                                                  & \textcolor{blue}{C}\ref{crit:ml_based}, \textcolor{blue}{C}\ref{crit:entity}                                        \\
					\quad QuickUMLS        & \cite{soldaini2016quickumls}                           & Python              & \cmark   & \cmark   & \cmark   & \cmark   & \cmark   & \cmark   & \cmark   & RB          & RB      &  2016  & 10/2023               & 215       & MIT license $\dagger$                                  & \textcolor{blue}{C}\ref{crit:ml_based}, \textcolor{blue}{C}\ref{crit:license}                                                                              \\
					\quad Bio-NLP          & \cite{badenes2022overview}                             & Python, Web         & \cmark   &          & \cmark   & \cmark   &          &          &          & N           & RB      &  2022  & 03/2022               & 2         & Apache-2.0 license                                     & \textcolor{blue}{C}\ref{crit:entity}                                                                                \\
					\midrule                                                                                                                                                                                                       
					\textbf{Selected}      &                                                        &                     &          &          &          &          &          &          &          &             &         &        &                       &           &                                                        &           \\                           
					\quad PubTator Central & \cite{weiPubTatorCentralAutomated2019}                 & REST/Tools          & \cmark   & \cmark   & \cmark   & \cmark   & \cmark   & \cmark   &          & ML / N      & RB      &  2019  & -                     & 315       & N/A                                                    &           \\                           
					\quad BERN2            & \cite{Bern2AnAdvanMujeen,kim2019neural}                & Python/Web          & \cmark   & \cmark   & \cmark   & \cmark   & \cmark   & \cmark   & \cmark   & N           & RB + N  &  2022  & 11/2023               & 46        & BSD 2-Clause "Simplified"                              &           \\                           
					\quad SciSpacy         & \cite{neumann2019scispacy}                             & Python              & (\cmark) & \cmark   & \cmark   & \cmark   & \cmark   & (\cmark) & \cmark   & N           & RB      &  2019  & 10/2023               & 635       & Apache 2.0                                             &           \\                           
					\quad bent             & \cite{LinkingChemicaRuas2020,ruas2023lasige}           & Python              & \cmark   & \cmark   & \cmark   & \cmark   & \cmark   & (\cmark) & \cmark   & N           & RB      &  2020  & 12/2023               & 13        & Apache 2.0                                             &           \\                           
					\quad HunFlair2    	   & \cite{weber2021hunflair}                               & Python              & \cmark   & \cmark   & \cmark   & \cmark   & \cmark   &          &          & N           & RB + N  &  2021  & 01/2024               & 83        & MIT License                                            &           \\                           
					\bottomrule
				\end{tabular}%
				\label{tab:tools_overview}%
			\end{adjustbox}
		\end{table}
	\end{landscape}
}

In Table~\ref{tab:tools_overview}, we present each tool considered for this study and their exclusion criterion if not
included.

	\section{Corpora}\label{app:corpora}



We briefly describe the datasets we use for the entity extraction evaluation.
\medskip

\paragraph{BioID}

The BioID corpus was created for the BioCreative VI shared task \enquote{Track 1: Interactive Bio-ID Assignment (IAT-ID)}.
Participants of the task were ask to annotate text originating from figure captions
with the entity types and IDs for organisms, genes, proteins, miRNA, small molecules, cellular components, cell types and cell lines, tissues and organs.
The corpus contains annotated captions for a total of 570 articles.
For NER, in our evaluation, we make use of the chemical, gene and organism (species) annotations given in the data. 
Note that we do not use the cell line annotations, because one of the selected tools (PubTator) uses these annotations during training and testing.
For entity extraction use only the species entity type, as other entities are normalized to KBs not supported by the tools selected.

\paragraph{tmVar (v3)}

The corpus contains 500 abstracts of PubMed articles manually annotated with different types of genetic variant and gene mentions.  
In total, the corpus contains over 4,000 mentions of genes as well as their NCBI gene identifiers.
In our study, we use the gene entity mentions both evaluations.


\paragraph{MedMentions}

The MedMentions corpus consists of 4,392 PubMed abstracts, which were randomly chosen from the papers published on PubMed in 2016.
The corpus provides annotations of entity mentions linked to UMLS spanning all its Semantic Types (entity types).
We use the the ST21pv (21 Semantic Types and Preferred Vocabularies) split.
For both NER and NEN, we evaluate on chemical and disease entities which we can be mapped to CTD identifiers via UMLS cross-reference tables.

        \section{Macro F1-score calculation}\label{app:macro_f1_score}

In practice, our macro F1-score calculation looks a bit different then presented in Section \ref{sec:entity_distribution}. Instead of calculating individual F1-scores for every database entity, we first compute individual precision and recall scores for each entity whenever possible. We then average those scores to obtain macro precision and macro recall and only then calculate their harmonic mean as our macro F1-score. This adapted calculation has a simple rationale: For rarely occurring entities in a given corpus, e.g., zebrafish (\enquote{NCBI taxon 7955}), there might often either be no recall or no precision score defined. In our zebrafish example, there might be a false positive prediction for zebrafish in a given corpus but neither true positives nor false negatives as the zebrafish entity might have never been part of the original corpus. This means there is no individual recall score to be defined for the zebrafish entity and thus no F1-score. However, we can still take into account the precision score defined for this entity (0 in this case) into the calculation of the macro precision score as a whole.

	\section{Comparison of HunFlair to HunFlair2}\label{app:hunflair_comparison}

\begin{table}[htbp]
  \centering
  \caption{Comparing F1-scores for NER in HunFlair versions 1 and 2.}
  \resizebox{\columnwidth}{!}{
    \begin{tabular}{lrrr}
\midrule    \textbf{Entity type / dataset} & \multicolumn{1}{l}{\textbf{HunFlair2}} & \multicolumn{1}{l}{\textbf{HunFlair-v1}} & \multicolumn{1}{l}{\textbf{Difference}} \\
    \midrule
    \textit{Chemical} &       &       &  \\
    MedMentions       & 58.40 & 52.78 & \textit{+5.62} \\
    \midrule
    \textit{Disease} &       &       &  \\
    MedMentions      & 62.18 & 63.01 & \textit{-0.83} \\
    \midrule
    \textit{Gene} &       &       &  \\
    tmVar (v3) & 87.87 & 83.21 & \textit{+4.66} \\
    \midrule
    \textit{Species} &       &       &  \\
    BioID & 58.21 & 57.62 & \textit{+0.59} \\
    \midrule
    Avg. All & 66.67 & 64.16 & \textit{+2.51} \\
    \bottomrule
    \end{tabular}%
    }
\label{tab:comparison_hunflair_versions}%
\end{table}%

In Table \ref{tab:comparison_hunflair_versions}, we report the comparison of results between HunFlair and HunFlair2. 
The disease entities are the only ones who do not profit from a joint NER model as they lose on average -1.57 pp on the F1 score compared to the disease-specific entity recognizer in HunFlair. 
Overall, HunFlair2 achieves a 2.02 pp improvement over HunFlair averaged over all entity types.
%
%

	\section{Named entity recognition}

\subsection{Named entity recognition: Evaluation}\label{app:evaluation_ner}

Following previous studies \cite{giorgi2020towards, weber2021hunflair} we report F1 scores comparing predicted spans to gold ones.
We classify a predicted span as a true positive (TP) if it either (i) matches exactly a gold span or
(ii) differs with the gold span by only one character either at  the beginning or the end. 
This is to account different handling of special characters by different tools which may result in minor span differences.
To ensure maximal fairness,
we have identified and removed all sentences (and documents) in the corpora used to train HunFlair2
which present an overlap with the corpora selected for our benchmarking.
Overlap is computed by matching all strings in each corpora on a sentence level. 
We note that the only significant overlap we found was between BioRED and tmVar v3 with 2,172 out of 6,755 sentences overlapping. 
For the evaluations of the SciSpacy on our corpora, we compute results using all its available NER models\footnote{See \url{https://allenai.github.io/scispacy/} for details.}
(en\_ner\_bc5cdr\_md, en\_ner\_bionlp13cg\_md, en\_ner\_craft\_md, en\_ner\_jnlpba\_md) and 
 report the result of the one that performs the best on the given corpora.
 Table \ref{tab:scispacy_ner_models} reports the combinations of SciScpacy model and corpus.
 
\begin{table*}[]
	\centering
        \caption{Best SciSpacy NER model on each of our evaluation corpora.}
	\begin{tabular}{c|c}
                \toprule
            \textbf{Dataset}                & \textbf{SciSpacy model}          \\ \midrule
		MedMentions (Chemical) & en\_ner\_bc5cdr\_md     \\
		MedMentions (Disease)  & en\_ner\_bc5cdr\_md     \\
		tmVar (v3)             & en\_ner\_bionlp13cg\_md \\
		BioID (Species)        & en\_ner\_craft\_md      \\
                \bottomrule
	\end{tabular}
	\label{tab:scispacy_ner_models}
\end{table*}

\subsection{Named entity recognition: Results}\label{app:ner_results}

\begin{table*}[tbhp]
	\centering
	\caption{Mention-level named entity recognition (NER) results evaluated by F1 scores.
		For each tool we report the performance differences between NER and the end-to-end entity normalization results in parenthesis.
		Bold figures highlight the highest value per row.
	}
	\begin{tabular}{l|ccccc}
		\toprule
		                  & \textbf{BERN2}                      & \textbf{HunFlair2} & \textbf{PubTator} & \textbf{SciSpacy} & \textbf{bent}     \\
		\midrule
		\textit{Chemical} &                                     &                    &                   &                   &                   \\
		\quad MedMentions & 50.81                               & \textbf{58.40}     & 38.53             & 43.98             & 52.20             \\
		                  & \textit{(+9.02/ +17.39$\dagger$)}   & \textit{(+7.23)}  & \textit{(+7.25)}  & \textit{(+9.03)}  & \textit{(+11.30)} \\

		\midrule
		\textit{Disease}  &                                     &                    &                   &                   &                   \\
		\quad MedMentions & 59.54                               & \textbf{62.18}              & 46.68             & 49.87             & 61.65    \\
		                  & \textit{(+12.21)}                   & \textit{(+4.61)}  & \textit{(+5.57)}  & \textit{(+9.09)}  & \textit{(+15.71)} \\

		\midrule
		\textit{Gene}     &                                     &                    &                   &                   &                   \\
		\quad tmVar (v3)  & 48.05                               & 87.87              & \textbf{88.82}    & 65.26             & 82.27             \\
		                  & \textit{(+4.09)}                    & \textit{(+11.12)}  & \textit{(+2.80)}   & -                 & \textit{(+81.73)} \\

		\midrule
		\textit{Species}  &                                     &                    &                   &                   &                   \\
		\quad BioID       & 60.80                               & 58.21              & 60.35             & 43.94             & \textbf{64.91}    \\
		                  & \textit{(+46.45)}                   & \textit{(+8.55)}   & \textit{(+1.45)}  & \textit{(+6.80)}   & \textit{(+54.56)} \\

		\midrule
		Avg. All          & 54.80                               & \textbf{66.67}     & 58.60             & 50.76             & 65.26             \\
		                  & \textit{(+17.94 / +21.38$\dagger$)} & \textit{(+7.88)}   & \textit{(+4.27)}  & \textit{(+13.15)} & \textit{(+40.83)} \\
		\bottomrule
	\end{tabular}%
	\label{tab:ner_results}%
\end{table*}%

\begin{table*}[tbhp]
	\centering
	\caption{Named entity recognition results using a lenient evaluation setting,
		i.e. we count each prediction as true positive which is a sub- or superstring of gold standard entity mention.
        For each tool we report the performance differences between the strict and lenient NER results in parenthesis.
  }
	\begin{tabular}{l|ccccc}
		\toprule

		                  & \textbf{BERN2}    & \textbf{HunFlair2} & \textbf{PubTator} & \textbf{SciSpacy} & \textbf{bent}    \\
		\midrule
		\textit{Chemical} &                   &                    &                   &                   &                  \\
		\quad MedMentions & 56.51             & \textbf{62.93}     & 44.87             & 48.66             & 57.51            \\
		                  & \textit{(+5.70)}  & \textit{(+4.53)}   & \textit{(+6.34)}  & \textit{(+4.68)}  & \textit{(+5.31)} \\
		\midrule
		\textit{Disease}  &                   &                    &                   &                   &                  \\
		\quad MedMentions & 67.20             & \textbf{69.47}              & 58.31             & 60.91             & 69.42   \\
		                  & \textit{(+7.66)}  & \textit{(+7.29)}   & \textit{(+11.63)} & \textit{(+11.04)} & \textit{(+7.77)} \\
		\midrule
		\textit{Gene}     &                   &                    &                   &                   &                  \\
		\quad tmVar (v3)  & 88.76             & \textbf{92.35}     & 89.46             & 74.44             & 89.48            \\
		                  & \textit{(+40.71)} & \textit{(+4.48)}   & \textit{(+0.64)}  & \textit{(+9.18)}  & \textit{(+7.21)} \\
		\midrule
		\textit{Species}  &                   &                    &                   &                   &                  \\
		\quad BioID       & \textbf{68.46}    & 64.99              & 62.31             & 48.39             & 67.82            \\
		                  & \textit{(+7.66)}  & \textit{(+6.78)}   & \textit{(+1.96)}  & \textit{(+4.45)}  & \textit{(+2.91)} \\
		\midrule
		Avg. All          & 70.23             & \textbf{72.44}     & 63.74             & 58.10             & 71.06            \\
		                  & \textit{(+15.43)} & \textit{(+5.77)}   & \textit{(+5.14)}  & \textit{(+7.34)}  & \textit{(+5.80)} \\
		\bottomrule
	\end{tabular}%
	\label{tab:lenient_ner_results}%
\end{table*}%

In Table \ref{tab:ner_results} we report Table \ref{tab:lenient_ner_results} the NER results by corpus for all the tools with standard (\nameref{sec:evaluation_nen}) and lenient (\nameref{results:ner}) evaluation, respectively.

\subsection{Named entity recognition: Infrequent entities}\label{app:infrequent_entities_ner}

\begin{table*}[tbhp]
	\centering
	\caption{Macro F1 scores for Named Entity Recognition over unique mentions. We compare the results to the micro-average F1 scores for NER in Table \ref{tab:ner_results}.}
	\begin{tabular}{l|cccccc}
		\toprule
		                  & \textbf{BERN2}    & \textbf{HunFlair2}     & \textbf{PubTator}      & \textbf{SciSpacy} & \textbf{bent}       \\
		\midrule
		\textit{Chemical} &                   &                        &                        &                   &                     \\
		\quad MedMentions & 33.43             & 38.56                  & 22.46                  & 25.35             & 35.43               \\
		                  & \textit{(-17.38)} & \textit{(-19.84)}      & \textit{(-16.07)}      & \textit{(-18.63)} & \textit{(-16.77)}   \\
		\midrule
		\textit{Disease}  &                   &                        &                        &                   &                     \\
		\quad MedMentions & 31.80             & 34.20                  & 21.02                  & 23.23             & 32.87               \\
		                  & \textit{(-27.74)} & \textit{(-27.98)}      & \textit{(-25.66)}      & \textit{(-26.64)} & \textit{(-28.78)}   \\
		\midrule
		\textit{Gene}     &                   &                        &                        &                   &                     \\
		\quad tmVar (v3)  & 26.88             & 66.74                  & 72.01                  & 38.42             & 58.42               \\
		                  & \textit{(-21.17)} & \textit{(-21.13)}      & \textit{(-16.81)}      & \textit{(-26.84)} & \textit{(-23.85)}   \\
		\midrule
		\textit{Species}  &                   &                        &                        &                   &                     \\
		\quad BioID       & 14.67             & 15.39                  & 16.67                  & 6.51              & 25.78             & \\
		                  & \textit{(-46.13)} & \textit{(-42.82)}      & \textit{(-43.68)}      & \textit{(-37.43)} & \textit{(-39.13)}   \\
		\midrule
    		Avg. All          & 26.70             & 38.72   & 33.04   & 23.38             & 38.13               \\
		                  & \textit{(-28.11)} & \textit{(-27.94)}      & \textit{(-25.56)}      & \textit{(-27.39)} & \textit{(-27.13)}   \\
		\bottomrule
	\end{tabular}%
	\label{tab:ner_results_entity_distribution}%
\end{table*}

Similarly to the infrequent entity analysis conducted for joint entity extraction in Section \nameref{sec:entity_distribution}, we conducted an analysis for infrequent entities for NER. The results can be found in Table \ref{tab:ner_results_entity_distribution}. We observe a consistent drop in F1-score performance of about 26 pp compared to the micro F1-score evaluation in Table \ref{tab:ner_results} across all tools. This shows that infrequent entities are not recognized as well as frequent ones indicating a need to further improve the capabilities of current biomedical NER models to handle those low-tail entities.

	\section{Named entity extraction results}

\begin{table*}
\centering
\caption{Cross-corpus micro-average precision, recall and F1 score of all tools evaluated tools.}\label{tab:prf1_nen}
\begin{tabular}{l|rrr|rrr|rrr|rrr|rrr}
	\toprule
	                  & \multicolumn{3}{c|}{\textbf{BERN2}}
	                  & \multicolumn{3}{c|}{\textbf{PubTator}}
	                  & \multicolumn{3}{c|}{\textbf{SciSpacy}}
	                  & \multicolumn{3}{c|}{\textbf{bent}}                                                                                                                      
	                  & \multicolumn{3}{c}{\textbf{HunFlair2}}\\
	                  & \textbf{P}
	                  & \textbf{R}
	                  & \textbf{F1}
	                  & \textbf{P}
	                  & \textbf{R}
	                  & \textbf{F1}
	                  & \textbf{P}
	                  & \textbf{R}
	                  & \textbf{F1}
	                  & \textbf{P}
	                  & \textbf{R}
	                  & \textbf{F1}
	                  & \textbf{P}
	                  & \textbf{R}
	                  & \textbf{F1}
	\\
	\midrule
	\textit{Chemical} &       &       &       &       &       &       &       &       &       &       &       &       &       &       &       \\
	\quad MedMentions & 48.08 & 36.95 & 41.79 & 39.81 & 25.77 & 31.28 & 41.40 & 30.24 & 34.95 & 46.76 & 36.35 & 40.90 & 52.32 & 50.08 & 51.17 \\
	\midrule
	\textit{Disease}  &       &       &       &       &       &       &       &       &       &       &       &       &       &       &       \\
	\quad MedMentions & 43.09 & 52.51 & 47.33 & 40.33 & 41.93 & 41.11 & 38.38 & 43.49 & 40.78 & 41.27 & 51.81 & 45.94 & 50.78 & 66.45 & 57.57 \\
	\midrule
	\textit{Gene}     &       &       &       &       &       &       &       &       &       &       &       &       &       &       &       \\
	\quad tmvar (v3)  & 40.37 & 48.26 & 43.96 & 90.33 & 82.11 & 86.02 & -  & -  & -  & 0.52  & 0.57  & 0.54  & 74.58 & 79.06 & 76.75 \\
	\midrule
	\textit{Species}  &       &       &       &       &       &       &       &       &       &       &       &       &       &       &       \\
	\quad BioID       & 17.93 & 11.96 & 14.35 & 81.52 & 46.11 & 58.90 & 50.66 & 29.31 & 37.14 & 11.79 & 9.22  & 10.35 & 65.94 & 39.83 & 49.66 \\
	\midrule
	Avg. All          & 37.37 & 37.42 & 36.86 & 62.99 & 48.98 & 54.33 & 43.48 & 34.01 & 37.61 & 25.08 & 24.49 & 24.43 & 60.90 & 58.85 & 58.79 \\
	\bottomrule
\end{tabular}
\end{table*}

In Table \ref{tab:prf1_nen} we report the cross-corpus micro-average precision, recall and F1 score of all evaluated tools.

\subsection{Document-level name entity extraction results}\label{app:nen_doc_level}
\begin{table*}[tbhp]
	\centering
	\caption{Cross-corpus evaluation results for \textit{document-level} named entity extraction.
		Results are micro-F1 (average over entities) score on the entire corpora.
		In parenthesis we report the difference with the \textit{mention-level} performance.
	}

	\label{tab:nen_results_dl}
	\begin{tabular}{l|ccccc}
	\toprule
	                 & \textbf{BERN2}    & \textbf{HunFlair2} & \textbf{PubTator} & \textbf{SciSpacy} & \textbf{bent}    \\
	\midrule
	\textit{Chemical} &                   &                    &                   &                   &                  \\
	\quad MedMentions & 41.86             & \textbf{48.93}     & 35.20             & 37.90             & 41.90            \\
	                & \textit{(+0.07)}  & \textit{(-2.24)}   & \textit{(+3.92)}  & \textit{(+2.95)}  & \textit{(+1.00)} \\
	\midrule
	\textit{Disease}  &                   &                    &                   &                   &                  \\
	\quad MedMentions & 51.35             & \textbf{64.58}     & 56.72             & 55.86             & 55.57            \\
	                 & \textit{(+4.02)}  & \textit{(+7.01)}  & \textit{(+15.67)} & \textit{(+15.08)} & \textit{(+9.63)} \\
	\midrule
	\textit{Gene}     &                   &                    &                   &                   &                  \\
	\quad tmVar v3    & 72.52             & 79.03              & \textbf{88.58}    & -                 & 0.64             \\
	                 & \textit{(+28.56)} & \textit{(+2.28)}   & \textit{(+2.56)}  & -                 & \textit{(+0.10)} \\
	\midrule
	\textit{Species}  &                   &                    &                   &                   &                  \\
	\quad BioID       & 16.38             & 49.49              & \textbf{54.50}    & 36.72             & 12.11            \\
	                 & \textit{(+2.03)}  & \textit{(-2.17)}   & \textit{(-4.40)}  & \textit{(-0.42)}  & \textit{(+1.76)} \\
	\midrule
	Avg               & 45.53             & \textbf{60.01}     & 58.75             & 43.15             & 27.56            \\
	                 & \textit{(+8.67)}  & \textit{(+1.22)}   & \textit{(+4.42)}  & \textit{(+5.87)}  & \textit{(+3.13)} \\
	\bottomrule
\end{tabular}

\end{table*}%

For a prediction to be correct the evaluation applied in Section \nameref{sec:results} requires
a match between both the mention boundaries and the normalization identifier with the gold standard.
This is a significantly challenging scenario, especially for a cross-corpus setting,
where different annotation guidelines impact what constitutes a mention.
To account for a more lenient type of evaluation, we report \textit{document-level} performance of all tools in Table \ref{tab:nen_results_dl}.
That is, we compare the set of unique gold standard and predicted identifiers gathered from all mentions of a document.
This type of evaluation also matches the use case of semantic indexing, a common application of entity extraction, in which all concepts mentioned in a document are indexed for improved information retrieval \cite{ChemicalIdentiLeaman2023}.

The results of the mention-level and document-level evaluation are not directly comparable, but it is shown that the tools can essentially achieve higher results.
One would expect that the document-level results would have to increase more for longer texts, since in these multiple mentions of the same entities occur likely more frequently and in a document-oriented evaluation the incorrect or missing identification of individual mentions is not taken into account.
Moreover, the recognition of exact mention boundaries doesn't impact the metric.
The considerations are confirmed by our results.
For instance, we see strong improvements for extraction of disease across all five tools.
The only exception is the recognition of species. In this case, better results can only be achieved for two of the five tools (BERN2 and bent).

        \section{Prediction overlaps}\label{app:overlap}

\begin{figure*}[tbhp]
\includegraphics[width=0.8\textwidth]{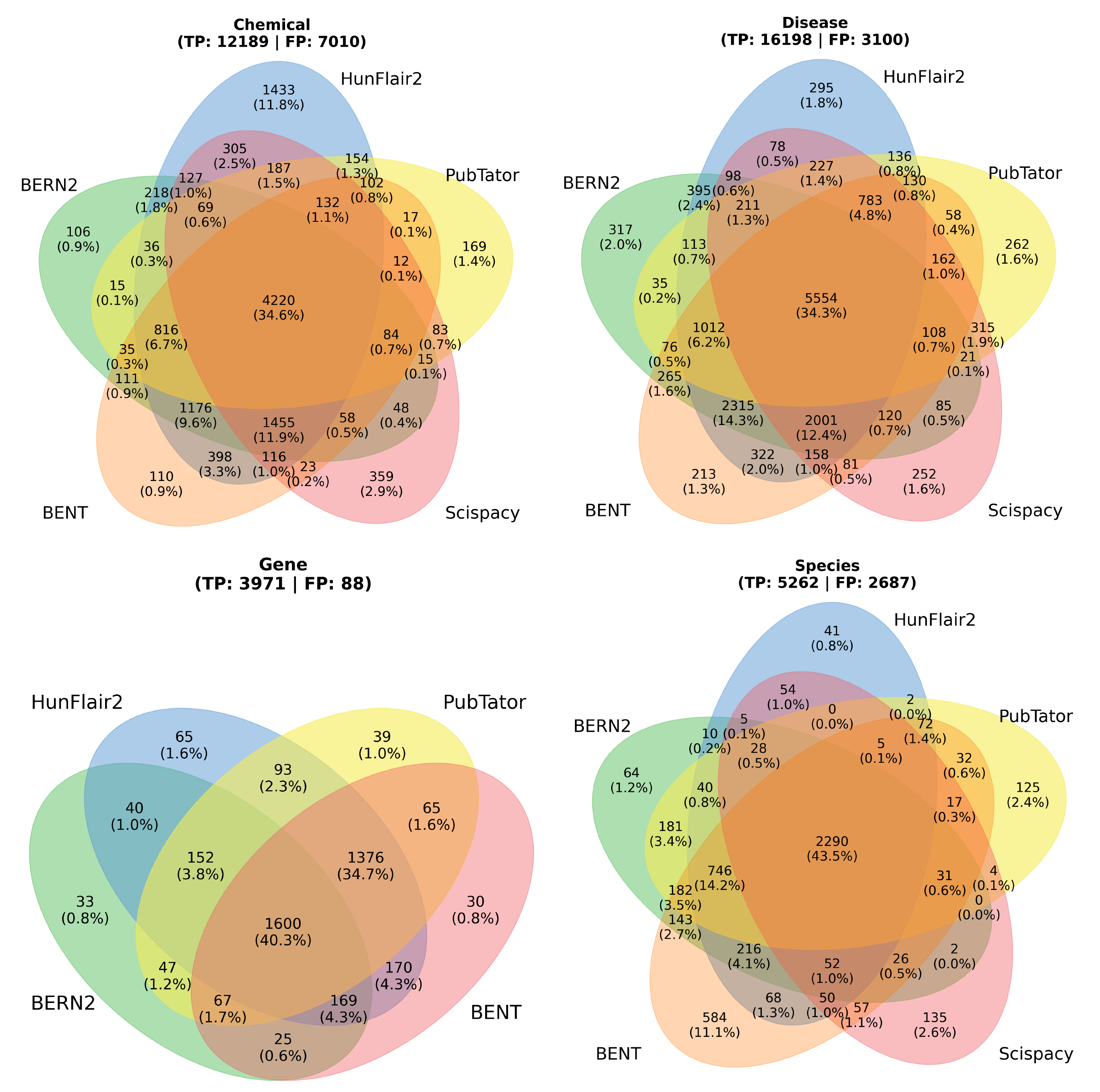}
\centering
\caption{Overview of the overlaps of the true positive predictions of BERN2, HunFlair2, PubTator, SciSpacy, and bent concerning the different entity types. 
For each setting, we report the total number of true positives (TP) found by at least one tool as well as the number of false positives (FP) in the sub-title.
For gene and tmVar (v3) we exclude Scispacy from the analysis as it does not support normalization of gene mentions.
}
\label{fig:tp_overlaps}
\end{figure*}
\end{appendices}

\end{document}